\documentclass[nohyperref]{article}

\usepackage{microtype}
\usepackage{graphicx}
\usepackage{subfigure}
\usepackage{booktabs} 

\usepackage{hyperref}

\usepackage[accepted]{icml2022}

\usepackage{amsmath}
\usepackage{amssymb}
\usepackage{mathtools}
\usepackage{amsthm}

\usepackage[capitalize,noabbrev]{cleveref}

\theoremstyle{plain}

\theoremstyle{definition}

\theoremstyle{remark}

\usepackage{siunitx}
\usepackage{latexsym}
\usepackage{times}
\usepackage{float}
\usepackage{footnote}
\usepackage{caption}
\usepackage{graphicx}
 \usepackage{booktabs}
 \usepackage{makecell}
 \usepackage{multirow}
\makesavenoteenv{tabular}
\makesavenoteenv{table}
\usepackage{graphicx}
\usepackage{hhline}
\usepackage{calc}
\usepackage{colortbl}
\usepackage{hyperref}
\usepackage{latexsym}
\usepackage{standalone}
\usepackage{url}
\usepackage{tikz}
\usepackage{soul}
\usepackage{fixltx2e}
\usepackage[normalem]{ulem}
\usepackage{color}
\usetikzlibrary{arrows,arrows.meta,decorations.pathmorphing,backgrounds,positioning,fit,petri,calc,math}
\usepackage{pgfmath}

\definecolor{light-gray}{gray}{0.85}

\DeclareRobustCommand{\best}[1]{{\sethlcolor{light-gray}\hl{#1}}}

\usepackage[textsize=tiny]{todonotes}

\icmltitlerunning{Unsupervised Detection of Contextualized Embedding Bias with Application to Ideology}

\begin{document}

\twocolumn[
\icmltitle{Unsupervised Detection of Contextualized Embedding\\Bias with Application to Ideology}



\icmlsetsymbol{equal}{*}

\begin{icmlauthorlist}
\icmlauthor{Valentin Hofmann}{ox_ling,lmu_cis}
\icmlauthor{Janet B. Pierrehumbert}{ox_eng,ox_ling}
\icmlauthor{Hinrich Schütze}{lmu_cis}
\end{icmlauthorlist}

\icmlaffiliation{ox_ling}{Faculty of Linguistics, University of Oxford}
\icmlaffiliation{ox_eng}{Department of Engineering Science, University of Oxford}
\icmlaffiliation{lmu_cis}{Center for Information and Language Processing, LMU Munich}

\icmlcorrespondingauthor{Valentin Hofmann}{valentin.hofmann@ling-phil.ox.ac.uk}

\icmlkeywords{Machine Learning, ICML}

\vskip 0.3in
]



\printAffiliationsAndNotice{}  

\begin{abstract}
We propose
a fully unsupervised method to detect bias
in contextualized embeddings. 
The method leverages the assortative information latently encoded 
by social networks and combines orthogonality regularization, structured
sparsity learning, and graph neural networks to 
find the embedding subspace capturing this information. As a concrete example, we focus on the phenomenon of ideological bias: we 
introduce the concept
of an ideological subspace, show how it can be found by applying our method to online discussion forums, 
and present techniques to 
probe it.
Our experiments suggest that the ideological subspace
encodes abstract evaluative semantics and reflects
changes in the political left-right spectrum during the presidency
of Donald Trump.
\end{abstract}

\section{Introduction}

What kinds of biases are implicitly encoded by word embeddings? 
This question has attracted considerable attention recently, with a
focus on gender \citep{Bolukbasi.2016, Caliskan.2017, Zhao.2019} and
race \citep{Tan.2019, Jiang.2020b, Guo.2021}. There has also been work on \textbf{ideological bias},
i.e., the association of word embeddings with political ideology resulting from framing \citep{Webson.2020, Bianchi.2021, Rozado.2021}.

Geometrically, bias can 
be represented as a \textbf{linear subspace} in embedding space that captures
most of the relevant semantic information \citep{Vargas.2020}. Prior studies have typically taken 
a supervised approach to detect this subspace, drawing upon external resources (e.g., word lists of gender-specific job titles). In the case of ideological bias, additional supervision is required
 to distinguish texts stemming from different ideologies (e.g., manual labels).
 The heavy need for supervision limits the scalability and wider applicability 
 of research on embedding bias.

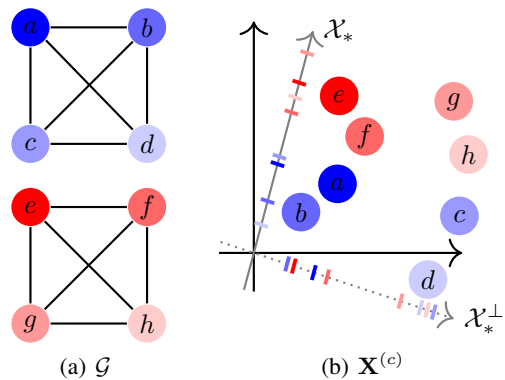
\begin{figure}
        \centering
        \begin{subfigure}[\small $\mathcal{G}$]{
            \centering
            \begin{tikzpicture}
[xscale=0.9,
 thick,
 vertex/.style={circle,
inner sep=0pt,minimum size=5mm}]

\node[vertex, fill=blue!60](s_1){$b$};
\node[vertex, fill=blue](s_2)[left=1cm of s_1] {$a$};
\node[vertex, fill=blue!20](s_3) [below=1cm of s_1] {$d$};
\node[vertex, fill=blue!40](s_4)[below=1cm of s_2]  {$c$};

\node[vertex, fill=red](s_5)[below=0.3cm of s_4]  {$e$};
\node[vertex, fill=red!60](s_6) [right=1cm of s_5] {$f$};
\node[vertex, fill=red!40](s_7) [below=1cm of s_5]  {$g$};
\node[vertex, fill=red!20](s_8) [below=1cm of s_6]  {$h$};

\draw[](s_1) to (s_2);
\draw[](s_1) to (s_3);
\draw[](s_1) to (s_4);
\draw[](s_2) to (s_3);
\draw[](s_2) to (s_4);
\draw[](s_3) to (s_4);

\draw[](s_5) to (s_6);
\draw[](s_5) to (s_7);
\draw[](s_5) to (s_8);
\draw[](s_6) to (s_7);
\draw[](s_6) to (s_8);
\draw[](s_7) to (s_8);

\end{tikzpicture}} 
  \label{fig:tutorial-1}
        \end{subfigure}
        \hspace{0.25cm}
        \begin{subfigure}[\small $\mathbf{X}^{(c)}$]{  
            \centering 
            \begin{tikzpicture}[xscale=0.9,
 thick,
 vertex/.style={circle,
inner sep=0pt,minimum size=5mm}]

\draw[-{>[scale=1.5]}] (-0.5,0) -- (3,0) node[anchor=north]{};
\draw[-{>[scale=1.5]}] (0,-0.5) -- (0,3) node[anchor=north]{};

\begin{scope}[rotate=-16.5]

\draw[-{>[scale=1.5]}, gray, dotted] (-0.5,0) -- (3,0)node[right, black] {$\mathcal{X}^\bot_*$};
\draw[-{>[scale=1.5]}, gray] (0,-0.5) -- (0,3) node[right, black] {$\mathcal{X}_*$};

\tikzmath{\x1 = 0.5; \y1 = 0.7; 
\x2 = 0.9; \y2 = 1.2; 
\x3 = 2.5; \y3 = 0.38; 
\x4 = 2.7; \y4 = 1.3; 
\x5 = 1.1; \y5 = 1.9; 
\x6 = 0.6; \y6 = 2.3; 
\x7 = 2.6; \y7 = 2.1; 
\x8 = 2.2; \y8 = 2.7; 
}

\node[vertex, fill=blue!60](b) at (\x1,\y1) {$b$};
\draw[blue!60,line width=0.5mm] (\x1,-0.1) -- (\x1,0.1) {};
\draw[blue!60,line width=0.5mm] (-0.1,\y1) -- (0.1,\y1) {};

\node[vertex, fill=blue](a) at (\x2,\y2) {$a$};
\draw[blue,line width=0.5mm] (\x2,-0.1) -- (\x2,0.1) {};
\draw[blue,line width=0.5mm] (-0.1,\y2) -- (0.1,\y2) {};

\node[vertex, fill=blue!20](d) at (\x3,\y3) {$d$};
\draw[blue!20,line width=0.5mm] (\x3,-0.1) -- (\x3,0.1) {};
\draw[blue!20,line width=0.5mm] (-0.1,\y3) -- (0.1,\y3) {};

\node[vertex, fill=blue!40](c) at (\x4,\y4) {$c$};
\draw[blue!40,line width=0.5mm] (\x4,-0.1) -- (\x4,0.1) {};
\draw[blue!40,line width=0.5mm] (-0.1,\y4) -- (0.1,\y4) {};

\node[vertex, fill=red!60](f) at (\x5,\y5) {$f$};
\draw[red!60,line width=0.5mm] (\x5,-0.1) -- (\x5,0.1) {};
\draw[red!60,line width=0.5mm] (-0.1,\y5) -- (0.1,\y5) {};

\node[vertex, fill=red](e) at (\x6,\y6) {$e$};
\draw[red,line width=0.5mm] (\x6,-0.1) -- (\x6,0.1) {};
\draw[red,line width=0.5mm] (-0.1,\y6) -- (0.1,\y6) {};

\node[vertex, fill=red!20](h) at (\x7,\y7) {$h$};
\draw[red!20,line width=0.5mm] (\x7,-0.1) -- (\x7,0.1) {};
\draw[red!20,line width=0.5mm] (-0.1,\y7) -- (0.1,\y7) {};

\node[vertex, fill=red!40](g) at (\x8,\y8) {$g$};
\draw[red!40,line width=0.5mm] (\x8,-0.1) -- (\x8,0.1) {};
\draw[red!40,line width=0.5mm] (-0.1,\y8) -- (0.1,\y8) {};

\end{scope}

\end{tikzpicture}}   
            \label{fig:tutorial-2}
        \end{subfigure}
        \caption[]{Framework. The example graph $\mathcal{G}$ consists of two components reflecting 
        different ideologies as indicated by node color. By projecting the corresponding embeddings $\mathbf{X}^{(c)}$ into the subspace 
        capturing $\mathcal{G}$'s polarization ($\mathcal{X}_*$), 
        we obtain representations of lower dimensionality that
still allow for perfect predictions of $\mathcal{G}$'s edges.} 
        \label{fig:tutorial}
\end{figure}

In this paper, we propose a fully unsupervised method 
to detect bias in contextualized word embeddings. 
We draw upon the fact that
\textbf{social networks} (a ubiquitous data structure) tend to be homophilous, i.e., 
neighboring nodes often have similar characteristics \citep{McPherson.2001}.
As a result, 
the structure of social networks latently encodes assortative information about variables relevant to bias, including gender \citep{Psylla.2017}, race \citep{DiPrete.2011}, 
and ideology \citep{Conover.2011}. Our method 
exploits this to detect bias
by rotating and shrinking the embedding space such 
that the resulting subspace is maximally informative about the social network topology (Figure \ref{fig:tutorial}).
Algorithmically, we combine graph neural networks with orthogonality regularization and structured sparsity learning.

As a concrete application, we focus on online discussion forums, specifically Reddit,
which can be modeled as networks with subforums as nodes 
and edges based on user overlap \citep{Olson.2015, Kumar.2018}.
We leverage the ideological information 
encoded by such networks to identify the ideological bias subspace.
The unsupervised nature of our method makes it challenging to interpret 
the found subspace. We present two methods for remedy:
\textbf{semantic probing}, which analyzes lexical-semantic regularities, 
and \textbf{indexical probing},
which aims to uncover the hidden ideological topology of the subspace.

Our \textbf{contributions} are as follows. We propose
a fully unsupervised method that exploits the structure of social 
networks to detect bias in contextualized embeddings, 
focusing on the use case of ideological bias in online discussion 
forums. Our method combines orthogonality regularization, structured
sparsity learning, and graph neural networks. 
We also present techniques to 
probe the found
subspace.
Our experiments show that the ideological subspace
encodes abstract evaluative semantics and reflects
changes in the left-right spectrum during the presidency
of Donald Trump.\footnote{We make our code available at \url{https://github.com/valentinhofmann/unsupervised_bias}.}

\section{Related Work}

A lot of research on \textbf{bias} in NLP (see \citet{Blodgett.2020} and \citet{Cao.2022} for reviews)
has focused on linear subspaces in word embedding 
space that contain 
information about categories such as gender \citep{Bolukbasi.2016, Caliskan.2017, Basta.2019, Gonen.2019}
and race \citep{Tan.2019, Jiang.2020b, Guo.2021}. There are also studies that 
measure word embedding associations to analyze differences between ideologies
\citep{Knoche.2019, Tripodi.2019, Xie.2019,  Webson.2020, Bianchi.2021, Rozado.2021, Walter.2021}.
Our work differs from these studies
in various ways:
(i) it is fully unsupervised, i.e., it does not need a key word list to 
locate the subspace, nor does it need information 
about the ideological orientation of texts; (ii) 
it does not make assumptions about the number of ideologies and can handle multidimensional ideological
spaces (e.g., different
ideologies on all nodes), which is theoretically more sound \citep{Heckman.1997};
(iii) it uses contextualized embeddings, thus obviating the need to fit separate embeddings for each ideology
(which is computationally infeasible in our setup).

Research on 
\textbf{ideological polarization}
in the computational social sciences \citep{Adamic.2005,Yardi.2010,Conover.2011, Guerra.2013, Himelboim.2013, Weber.2013, Mejova.2014, Bakshy.2015, Garcia.2015,Sylwester.2015, Garimella.2018, Green.2020, Cann.2021, Waller.2021}
and NLP
\citep{Sagi.2013, Iyyer.2014, PreotiucPietro.2017, An.2018, Kulkarni.2018, An.2019, Demszky.2019, Shen.2019, Davoodi.2020, Mokhberian.2020, Roy.2020, Tyagi.2020, Vorakitphan.2020, He.2021, Mendelsohn.2021}
has shown that polarization can manifest itself on the
level of social networks
by a polarized network structure and on the level of
political discourse by a range of linguistic phenomena including framing.
Most closely related,
\citet{Hofmann.2022b}
leverage the structure of social networks
to detect polarized issues. Our work 
differs in both its topic and its metholodology: (i) it focuses on ideological bias 
and the embedding subspace containing it; (ii) it is fully unsupervised, which increases 
its applicability.

\section{Framing and Ideological Bias} \label{sec:theory}

Framing describes the mechanism by which 
proponents of different ideologies highlight different aspects of the same issue
during political communication, thereby lending greater perceived importance 
to them \citep{Entman.1993,  Nelson.1997, Druckman.2001, Chong.2007}.
In the US, e.g., liberals tend to frame immigrants as victims, underscoring their vulnerability,
whereas conservatives often frame them as criminals, portraying them as threats
to the public \citep{Benson.2013, Mendelsohn.2021}.

What is the relationship between framing and ideological bias in contextualized word embeddings?
Framing results in language-internal and language-external associations 
that interact in creating ideological bias. Linguistically,  
framing is realized by bringing certain words 
into syntactic contiguity with each other, impacting 
cooccurrence statistics and leading to 
\textbf{(first-order) semantic associations} (e.g., between \textit{immigrants} and \textit{criminals}).
Contextualized word embeddings encode such semantic associations by mapping words 
to vectors that vary with the context \citep{Coenen.2019, Field.2019, Wiedemann.2019},
placing, e.g., the embedding of \textit{immigrants} close to the embedding of \textit{criminals}.
Extralinguistically, 
certain frames are preferentially employed by proponents of certain ideologies, creating \textbf{(second-order) indexical associations}
\citep{Silverstein.2003, Nguyen.2021} between 
 the linguistic manifestations of framing and ideologies (e.g., between the semantic association of \textit{immigrants} with \textit{criminals} on the one hand and conservatives on the other).\footnote{Notice that while political ideology 
 is not typically viewed as a sociolinguistic variable \citep{Eckert.2012, Eckert.2019}, it impacts social identity construction in a similarly crucial way.} 
 Such indexical associations are reflected by systematic covariation 
 in the region occupied by the embeddings of a word and the ideological orientation 
 of the text on which the embeddings are computed, making it possible (in the extreme case) to 
 predict the ideology from the word embedding (e.g., predict that a text is conservative
 based on the fact that the embeddings of \textit{immigrants} are close to the embeddings of \textit{criminals}).

Our conceptualization of ideological bias is inherently neutral: 
 rather than examining its potentially harmful effects (a topic in its own right),
 we aim 
 to present ideological bias as a little-investigated property of contextualized word embeddings
 that can be used as an analytical lens to draw inferences about political reasoning.
 This focus sets our work apart from many other studies on bias in NLP \citep{Blodgett.2020, Cao.2022} and
 puts it more in line with research on the linguistic manifestations
 of political slant \citep{Gentzkow.2010, Fulgoni.2016, Fan.2019, Baly.2020}, which so far has not 
touched on the topology of contextualized word embeddings.

It is important to notice that the connection between social groups, 
the typical word cooccurrence patterns they employ, and the resulting associations 
in contextualized embedding space apply to many other types of bias as well (e.g., bias in the way words are used by people 
of different gender or bias in the way concepts are used by different scientific fields).
In all these cases,
 bias is only one factor besides many others 
 such as syntax \citep{Goldberg.2019, Hewitt.2019} impacting variation in the contextualized embeddings
 of a word. One of the key goals of this paper is to devise a method that 
 overcomes this challenge and separates, for a set of words, the variation in embedding space
 caused by bias from the variation caused by other factors.

\section{Ideological Subspace} \label{sec:contested}

Let $\mathcal{X} \subset \mathbb{R}^d$ be a $d$-dimensional embedding space.
We want to find the $d_*$-dimensional orthogonal subspace
$\mathcal{X}_* \subset \mathbb{R}^{d_*}$, with $d_* \ll d$,
that contains
all and only
information relevant to ideological framing, and whose orthogonal complement $\mathcal{X}^\bot_*$ 
contains information irrelevant to ideological framing.
We call 
$\mathcal{X}_*$ the \textbf{ideological subspace} of $\mathcal{X}$.

In this paper, we show how $\mathcal{X}_*$ can be found for
discussion forums that are
divided into smaller subforums. While 
there is typically no explicit information about the subforum ideologies that 
would allow us to perform supervised learning,
we argue that the network structure of the subforums 
is sufficient to determine $\mathcal{X}_*$.
More formally, let $\mathcal{G} = (\mathcal{V}, \mathcal{E})$ be a graph consisting of a set of 
subforums $\mathcal{V}$ and a set of edges between the
subforums $\mathcal{E}$ representing homophilous \citep{McPherson.2001} relations such as user overlap. 
Let $C$ be
a set of political concepts to be analyzed
and $X = \{ \mathbf{X}^{(c)} \}_{c \in C}$ a set of matrices with
\begin{equation}
\mathbf{X}^{(c)} = [ \mathbf{x}^{(c)}_1, \dots \mathbf{x}^{(c)}_{|\mathcal{V}|}]^\top,
\end{equation}
i.e., each row in  $\mathbf{X}^{(c)} \in \mathbb{R}^{|\mathcal{V}| \times d}$ contains the embedding $\mathbf{x}^{(c)}_i \in \mathcal{X}$ of concept $c$ for subforum $i$.
The embeddings capture the ideological framing of concept $c$ in subforum $i$ (which we want 
to be in $\mathcal{X}_*$) as well as other information irrelevant to ideology (which we want to be in $\mathcal{X}^\bot_*$).

Our key idea is that due to the homophily of $\mathcal{G}$, subforums close to (far from) each other 
in $\mathcal{G}$ are expected 
to be ideologically similar (dissimilar), which should be reflected 
by similar (dissimilar) patterns of ideological bias while
having little
effect on
other semantic characteristics. Put differently,
for $\mathbf{X}^{(c)}$ representing concept $c$, its
projection to $\mathcal{X}_*$ should be informative
about the proximity of two subforums in $\mathcal{G}$, but
its projection to $\mathcal{X}^\bot_*$ should not.
We formalize this idea as the task of predicting links in $\mathcal{G}$
using the embedding matrices in $X$ as features while at the same time shrinking 
$\mathcal{X}$ to $\mathcal{X}_*$, i.e., we leverage the training 
signal from link prediction to remove the task-irrelevant information in $\mathcal{X}^\bot_*$.

To make this more concrete,
Figure \ref{fig:tutorial} shows an example graph of eight nodes
that fall into two components reflecting distinct
ideologies
as well as the corresponding embeddings for one example concept.
By projecting the embeddings into the subspace capturing the network polarization ($\mathcal{X}_*$), 
we obtain representations that are of lower dimensionality while
still allowing for perfect predictions of the links in $\mathcal{G}$. Projecting into
the orthogonal complement ($\mathcal{X}^\bot_*$), on the other hand, does not allow for perfect predictions.

As a result of the two-level structure 
of ideological bias, 
$\mathcal{X}_*$ encodes both semantic and indexical information.
For the running example of immigrants, $\mathcal{X}_*$
might encode the semantic category of agency to capture the different framing as 
victims or criminals, and it might exhibit regions indexically linked to 
liberals and conservatives. This makes it possible to analze $\mathcal{X}_*$ from two complementary 
perspectives.

\section{Model} \label{sec:model}

We use pretrained (base, uncased) BERT \citep{Devlin.2019} to obtain subforum-specific representations 
for the concepts ($d$ is 768).\footnote{We take the mean-pooled embedding
if a concept is split into multiple WordPiece tokens.} Specifically, for all concepts $c \in C$ and subforums $i \in \mathcal{V}$, 
we compute average contextualized embeddings $\mathbf{x}^{(c)}_i$ 
and use them as 
node features
in a graph auto-encoder to perform link prediction on $\mathcal{G}$ (i.e., we predict edges between subforums).
Based on the assumption that the ideological subspace contains the information
needed for
this task, we simultaneously rotate and
shrink the
space to find the ideological subspace $\mathcal{X}_*$.

The first part of the model rotates $\mathbf{X}^{(c)}$ such that 
the information relevant to ideological bias corresponds to 
a small number of dimensions 
in the rotated space $\mathcal{X}_r$, i.e.,
\begin{equation}
 \mathbf{X}^{(c)}_r =  \mathbf{X}^{(c)} \mathbf{R}.
\end{equation}
Here,
$\mathbf{R} \in \mathbb{R}^{d \times d}$ is an orthogonal matrix that transforms $\mathcal{X}$ into $\mathcal{X}_r$. By choosing $\mathbf{R}$
to be orthogonal, we do not add or remove any information from 
the original space. $\mathbf{R}$ is optimized as part of the training. To enforce the orthogonality of $\mathbf{R}$, 
we use orthogonality regularization \citep{Bousmalis.2016, Brock.2017, Vorontsov.2017}, i.e., we 
compute an orthogonality penalty,
\begin{equation}
\mathcal{L}_o =  \| \mathbf{R}\mathbf{R}^\top - \mathbf{I} \|_F^2,
\end{equation}
where $\mathbf{I} \in \mathbb{R}^{d \times d}$ is 
the identity matrix, and $\| \cdot \|^2_F$ is the squared Frobenius norm.

To perform 
link prediction,
we use a graph auto-encoder with two convolutional layers \citep{Kipf.2016, Kipf.2017} 
that takes as input the rotated embeddings $\mathbf{X}^{(c)}_r$
as well as $\mathcal{G}$'s adjacency matrix $\mathbf{A} \in \mathbb{R}^{|\mathcal{V}| \times |\mathcal{V}|}$
and in each layer updates the embeddings according to
\begin{equation}
\mathbf{H}^{(l+1)} = \sigma \left( \tilde{\mathbf{D}}^{-\frac{1}{2}} \tilde{\mathbf{A}}
\tilde{\mathbf{D}}^{-\frac{1}{2}} \mathbf{H}^{(l)} \mathbf{W}^{(l)}
  \right),
\end{equation}
where $\mathbf{H}^{(l)}$ is the embedding matrix after layer $l$, $\tilde{\mathbf{A}} = \mathbf{A} + \mathbf{I}$ is $\mathcal{G}$'s adjacency matrix with added
self-loops, $\tilde{\mathbf{D}}$ is the degree matrix of $\tilde{\mathbf{A}}$, 
and $\mathbf{W}^{(l)}$ is the weight matrix of layer $l$. $\sigma$ is the activation function, 
for which we use ReLU after the first
and no non-linearity after the second layer. 
We set $\mathbf{H}^{(0)} = \mathbf{X}^{(c)}_r$.
We reconstruct $\mathbf{A}$
by means of a dot-product decoder \citep{Kipf.2016} and compute
a prediction loss $\mathcal{L}_p$ using binary cross-entropy.

\begin{table} [t!]
\caption{Dataset statistics. $|\mathcal{D}|$: number of comments; $|\mathcal{V}|$: number of nodes (subreddits); $|\mathcal{E}|$: number of 
edges; $\mu_d$: average node degree; $\mu_\pi$: average shortest path length; $\rho$: density; $Q$: maximum modularity.}  \label{tab:data-stats}
\centering
\resizebox{\linewidth}{!}{%
\begin{tabular}{@{}lrrrrrrr@{}}
\toprule
Year & $|\mathcal{D}|$ & $|\mathcal{V}|$ & $|\mathcal{E}|$ &  $\mu_d$  & $\mu_\pi$ & $\rho$ & $Q$\\
\midrule
2013 &6,306,458 & 108 & 324  & 6.00 & 3.08 & .056 & .560 \\
2014 &6,664,567 & 132 & 335  & 5.08 & 3.86 & .039 & .663 \\
2015 &9,230,022 & 168 & 493 & 5.87 & 3.87 & .035 & .672 \\
2016 &34,801,075 & 255 & 1,318  & 10.34 & 3.14 & .041 & .603  \\
2017 &38,278,685 & 295 & 1,572  & 10.66 & 3.14 & .036 & .585 \\
2018 & 40,222,627& 316 & 1,604 & 10.15 & 3.17 & .032 & .584 \\
2019 &46,590,000 & 412 & 2,536 & 12.31 & 3.20 & .030 & .603\\
\bottomrule
\end{tabular}}
\end{table}

To shrink $\mathcal{X}_r$, we combine the graph auto-encoder with structured sparsity learning \citep{Yuan.2006, Liu.2015,Lebedev.2016, Wen.2016, Yoon.2017}, which has the effect that entire rows of the weight matrix $\mathbf{W}^{(0)}$ are set to zero during training. Writing $\mathbf{W}^{(0)} 
= [  \mathbf{w}^{(0)}_1, \dots, 
\mathbf{w}^{(0)}_d ]^\top $ as a series of row vectors, 
we define the sparsity penalty as
\begin{equation}
\mathcal{L}_s = \sum_{j=1}^{d} \| \mathbf{w}^{(0)}_j \|_2.
\end{equation}
This is a mixed $\ell_1$/$\ell_2$ regularization (the $\ell_1$ norm of the
row $\ell_2$ norms) that leads to sparsity on the level of rows. When all 
entries in a row $\mathbf{w}^{(0)}_j$ are zero, this has the effect of
essentially removing dimension $j$ from $\mathcal{X}_r$. The rows with non-zero weights after training
determine the $d_*$ dimensions of the ideological subspace $\mathcal{X}_*$.

The final loss is
$\mathcal{L}
= \mathcal{L}_p + \lambda_o \mathcal{L}_o + \lambda_s \mathcal{L}_s$, 
where $\lambda_o, \lambda_s > 0$ are hyperparameters. Since 
$ \mathcal{L}_s$ is non-differentiable, we use
proximal gradient descent \citep{Bach.2011, Parikh.2013} for optimization.
 We
approximate the 
weighted proximal operator of the $\ell_1$/$\ell_2$ norm
using the Newton-Raphson algorithm \citep{Deleu.2021}.

\section{Experiments}

\subsection{Data}

We base our study on the Reddit Politosphere \citep{Hofmann.2022}, a dataset covering
 the political discourse on the social media platform Reddit from 2008 to 2019. For each year, the Reddit Politosphere
contains (i) a graph with political subforums (called \textit{subreddits} in the context of Reddit) as nodes and edges computed by applying
statistical backboning to the counts of users shared between subreddits, and (ii) all English comments posted in each of the political subreddits. To make training robust, we confine ourselves 
to years in which the graph has at least 100 nodes (2013 to 2019). Table \ref{tab:data-stats} gives summary statistics. 
As indicated by the high modularity values,
the graphs are polarized (nodes cluster by ideology).
For $C$, we draw upon year-wise lists of 1,000 English political concepts (unigrams such as \textit{abortion} and bigrams such as \textit{social security})
determined by comparing the vocabulary of the political subreddits with the vocabulary 
of the default subreddits (i.e., subreddits users used to be subscribed to automatically) 
using mutual information \citep{Hofmann.2022b}.

\subsection{Experimental Setup}

\begin{table*} [t!]
\caption{Performance on link prediction (MAUC). The embeddings in $\mathcal{X}_*$ perform similarly to the embeddings in $\mathcal{X}$ while being of much lower
  dimensionality. $d_*$ for $\mathcal{X}_*$ is 17 (2013, 2014, 2015), 13 (2016), 6 (2017), 19 (2018), and 6 (2019) versus 768 for $\mathcal{X}$. Differences between better (gray) and lower performance are significant (underlined) for only six
  columns as shown by two-tailed $t$-tests ($p < .01$), underscoring the similar performance of $\mathcal{X}$ and $\mathcal{X}_*$.}  \label{tab:performance}
\centering
\resizebox{\linewidth}{!}{%
\begin{tabular}{@{}lrrrrrrrrrrrrrrrr@{}}
\toprule
{} & \multicolumn{8}{c}{Dev} & \multicolumn{8}{c}{ Test } \\
\cmidrule(lr){2-9}
\cmidrule(l){10-17}
Space & 2013 & 2014 & 2015 & 2016 & 2017 & 2018 & 2019 & $\mu\pm\sigma$ & 
2013 & 2014 & 2015 & 2016 & 2017 & 2018 & 2019 & $\mu\pm\sigma$\\
\midrule
$\mathcal{X}_*$ & \best{{\underline{.671}}} & \best{.717} & .656 & \best{.709} & \best{.700} & .657 & .728 & .691$\pm$.028 & \best{.699} & \best{.735} & .660 & .669 & \best{.687} & .695 & .696 & .692$\pm$.022\\ 
$\mathcal{X}$ & .649 & .699 & \best{{\underline{.681}}} & .705 & .695 & \best{{\underline{.673}}} & \best{{\underline{.741}}} & .692$\pm$.027 & .683 & .725 & \best{{\underline{.699}}} & \best{.683} & \best{.687} & \best{.700} & \best{{\underline{.707}}} & .698$\pm$.014\\
\bottomrule
\end{tabular}}
\end{table*}

To  estimate how much ideological information we lose by shrinking the space,
we compare against a model that directly uses the concept embeddings
as input to the graph auto-encoder, i.e., it neither rotates
nor shrinks the space.
Of course, 
the $\mathcal{X}_*$ embeddings will not improve over the 
$\mathcal{X}$ embeddings, but their performance
should not be substantially worse, i.e.,  the comparison tells us how much task-relevant information
we lose by removing $\mathcal{X}_*^\bot$.

We split concepts and edges for each year into train (60\%), dev (20\%), and test (20\%). Models are trained
separately for the years. Our training regime
consists of \textit{superepochs} in which we loop over all train concepts, and \textit{epochs} in which we predict the train edges between 
the subreddits 
using the embeddings of a certain concept as node features. Put differently, on each epoch 
we perform one pass through the model as described in Section~\ref{sec:model}, 
with the node features changing on every epoch as we loop over concepts. This is repeated for a chosen number of superepochs (i.e., if 
we train for 100 superepochs, the embeddings of each train concept are used 100 times as node features).
For evaluation, we loop over the dev (or test) concepts and compute the area under the curve (AUC)
that results from predicting the dev (or test) edges between the subreddits
using the embeddings of a certain concept as node features. Finally, we compute the mean AUC (MAUC) for all dev (or test) concepts.

We use 
gradient accumulation to make training robust, i.e., 
weights are only updated after 10 concepts (epochs in our training regime). 
We perform grid search for the learning rate $r \in \{ \num{1e-4}, \num{3e-4}, \num{1e-3} \}$. For the model
used to find $\mathcal{X}_*$, we further perform grid search for
the orthogonality constant $\lambda_o \in  \{ \num{1e-3}, \num{3e-3}, \num{1e-2} \}$ 
as well as the sparsity constant $\lambda_s \in  \{ \num{1e-2}, \num{3e-2}, \num{1e-1} \}$.
In total, there are 3 hyperparameter search trials for $\mathcal{X}$ and 27 for $\mathcal{X}_*$ per year.
We use Adam \citep{Kingma.2015} as the optimizer. See Appendix \ref{ap:setup} for further details about hyperparameter tuning and runtime.

\subsection{Performance}

\begin{figure}[t!]
        \centering
                \includegraphics[width=0.45\textwidth]{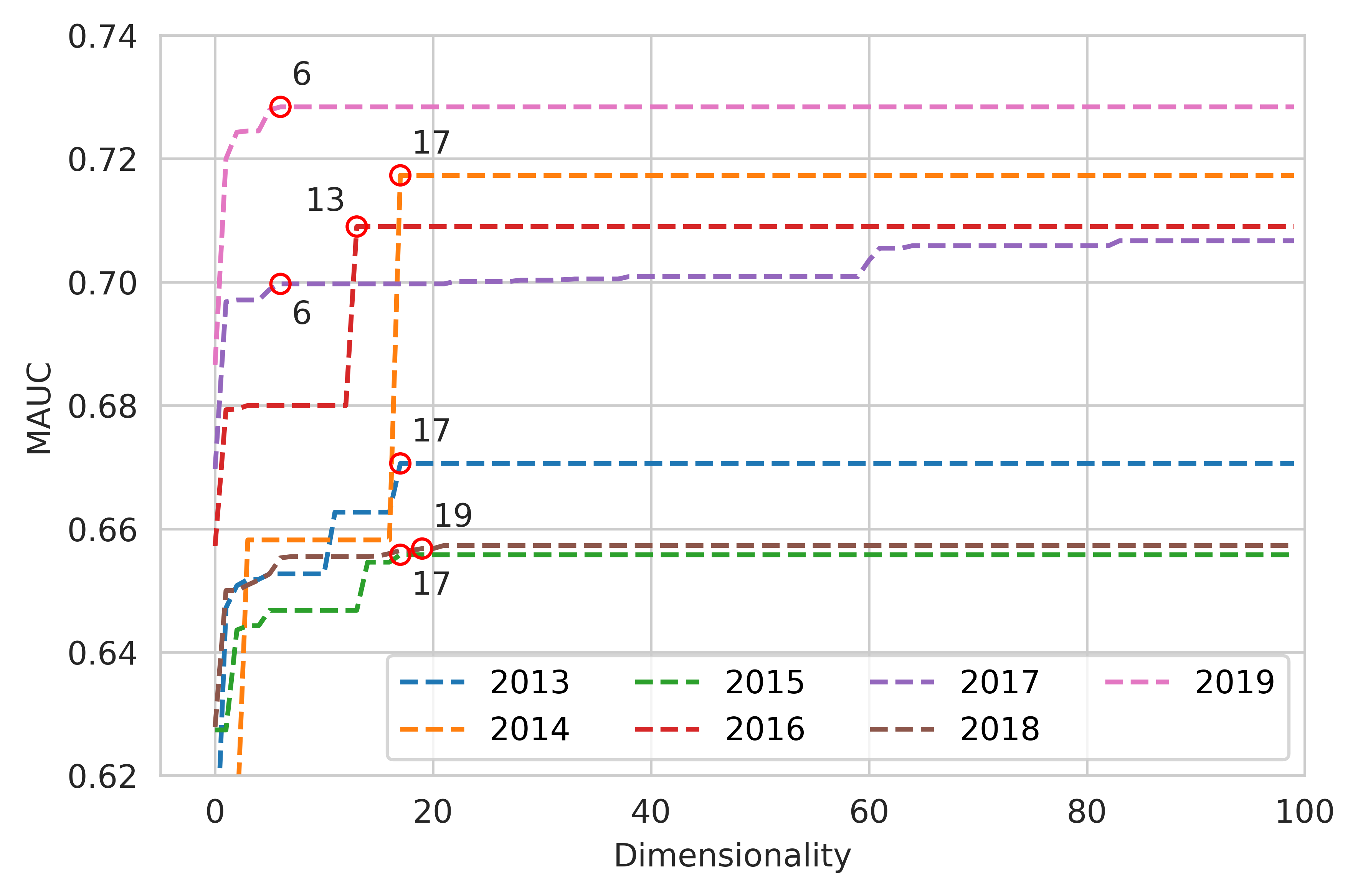}  
        \caption[]{Performance on link prediction (MAUC). The figure shows how the performance varies as a function of $d_*$, 
        the dimensionality of $\mathcal{X}_*$. We truncate at $d_* = 100$ since there is no further change for larger values. We
        highlight the found knees and corresponding values of $d_*$.}
        \label{fig:performance}
\end{figure}

The performance on link prediction as a function 
of $d_*$ exhibits a pronounced knee shape (Figure~\ref{fig:performance}), i.e., 
we can shrink $\mathcal{X}$ substantially
without losing
performance.
To detect the knee,
we use the algorithm proposed by \citet{Satopaa.2011}.
Table~\ref{tab:performance} shows that
when comparing to the baseline without 
sparsity constraint, the embeddings in
$\mathcal{X}_*$ (with tuned $d_*$) and $\mathcal{X}$ 
perform very similarly, i.e., the embeddings 
from the subspace allow to reconstruct the edges between the subreddits
with nearly identical performance as the embeddings from the full space. The difference in performance is statistically
insignificant for the majority of cases.
The fact that between 97.5\% (2018) and 99.2\% (2017 and 2019)
of $\mathcal{X}$ can be discarded 
 without major detrimental effects on performance suggests that 
 most of the information encoded by $\mathcal{X}$
 does not bear relevance for framing. This is in line with prior work showing that contextualized embeddings 
 represent various kinds of information such as syntax \citep{Goldberg.2019, Hewitt.2019} that do not play a role in framing.
 Interestingly, even though training is performed separately for the years, 
it converges to subspaces of similar sizes 
($5 < d_* < 20$), suggesting that the type of information encoded by 
$\mathcal{X}_*$ might also be similar.

What information is encoded by $\mathcal{X}_*$? Given 
the two-level structure of ideological bias, we examine this question
with a view on $\mathcal{X}_*$'s semantic and indexical associations.

\subsection{Semantic Probing of $\mathcal{X}_*$} \label{sec:semantic}

\begin{table*} [t!]
\caption{Concreteness and morality ratings for the 100 semantic axes with maximum and minimum $s_a$ scores. Semantic axes that are strongly encoded by $\mathcal{X}_*$ have consistently lower concreteness and higher morality scores than semantic axes that are weakly encoded by $\mathcal{X}_*$. The higher value per column (gray) is underlined if it is significantly ($p < .01$) higher than the lower value as
shown by a two-tailed $t$-test ($p < .01$).}  \label{tab:concreteness}
\centering
\resizebox{\linewidth}{!}{%
\begin{tabular}{@{}lrrrrrrrrrrrrrrrr@{}}
\toprule
{} & \multicolumn{8}{c}{Concreteness} & \multicolumn{8}{c}{ Morality } \\
\cmidrule(lr){2-9}
\cmidrule(l){10-17}
$s_a$ & 2013 & 2014 & 2015 & 2016 & 2017 & 2018 & 2019 & $\mu\pm\sigma$ & 
2013 & 2014 & 2015 & 2016 & 2017 & 2018 & 2019 & $\mu\pm\sigma$\\
\midrule
Max & .262 & .307 & .298 & .301 & .292 & .283 & .286 & .290$\pm$.014 & \best{.116} & \best{.126} & \best{{\underline{.128}}} & \best{{\underline{.124}}} & \best{.118} & \best{{\underline{.123}}} & \best{{\underline{.122}}} & .122$\pm$.004\\ 
Min & \best{{\underline{.423}}} & \best{{\underline{.389}}} & \best{{\underline{.443}}} & \best{{\underline{.487}}} & \best{{\underline{.383}}} & \best{{\underline{.370}}} & \best{{\underline{.405}}} & .414$\pm$.037 & .113 & .111 & .096 & .100 & .106 & .100 & .099 & .104$\pm$.006\\ 
\bottomrule
\end{tabular}}
\end{table*}

\begin{table} [t]
\caption{Top and bottom adjectival semantic axes. For each year, the table shows the four adjectival semantic axes with 
highest and lowest $s_a$ scores. While the top axes tend to have abstract evaluative meanings, this is 
not the case for bottom axes. Corresponding tables for nominal and verbal semantic axes are provided in Appendix \ref{ap:axes}.}  \label{tab:axes}
\centering
\resizebox{\linewidth}{!}{%
\begin{tabular}{@{}lllll}
\toprule
Year & Max $s_a$ & Min $s_a$ \\
\midrule
\multirow{4}{*} {2013} & \textit{executive/legislative} & \textit{aware/unaware}\\ 
 & \textit{immoral/moral} & \textit{adjacent/separate}\\ 
 & \textit{general/particular} & \textit{happy/unhappy}\\ 
 & \textit{autocratic/democratic} & \textit{cold/warm}\\ 
\midrule 
\multirow{4}{*} {2014} & \textit{constitutional/unconstitutional} & \textit{official/unofficial}\\ 
 & \textit{nonpartisan/partisan} & \textit{less/more}\\ 
 & \textit{capable/incapable} & \textit{first/second}\\ 
 & \textit{armed/unarmed} & \textit{primary/secondary}\\ 
\midrule 
\multirow{4}{*} {2015} & \textit{accurate/inaccurate} & \textit{first/second}\\ 
 & \textit{boring/interesting} & \textit{single/triple}\\ 
 & \textit{difficult/easy} & \textit{primary/secondary}\\ 
 & \textit{biased/impartial} & \textit{following/leading}\\ 
\midrule 
\multirow{4}{*} {2016} & \textit{useful/useless} & \textit{north/south}\\ 
 & \textit{ill/well} & \textit{following/leading}\\ 
 & \textit{expensive/inexpensive} & \textit{minus/plus}\\ 
 & \textit{common/uncommon} & \textit{dark/light}\\ 
\midrule 
\multirow{4}{*} {2017} & \textit{critical/uncritical} & \textit{former/latter}\\ 
 & \textit{central/peripheral} & \textit{happy/unhappy}\\ 
 & \textit{autocratic/democratic} & \textit{likely/unlikely}\\ 
 & \textit{scientific/unscientific} & \textit{aware/unaware}\\ 
\midrule 
\multirow{4}{*} {2018} & \textit{autocratic/democratic} & \textit{first/second}\\ 
 & \textit{critical/uncritical} & \textit{former/latter}\\ 
 & \textit{biased/impartial} & \textit{early/late}\\ 
 & \textit{armed/unarmed} & \textit{likely/unlikely}\\ 
\midrule 
\multirow{4}{*} {2019} & \textit{autocratic/democratic} & \textit{likely/unlikely}\\ 
 & \textit{national/transnational} & \textit{cold/warm}\\ 
 & \textit{biased/impartial} & \textit{different/similar}\\ 
 & \textit{qualified/unqualified} & \textit{former/latter}\\ 

\bottomrule
\end{tabular}}
\end{table}

\begin{figure}[t!]
        \centering           
        \begin{subfigure}[\small $\mathcal{X}_*$ (2013)]{  
            \includegraphics[width=0.11\textwidth]{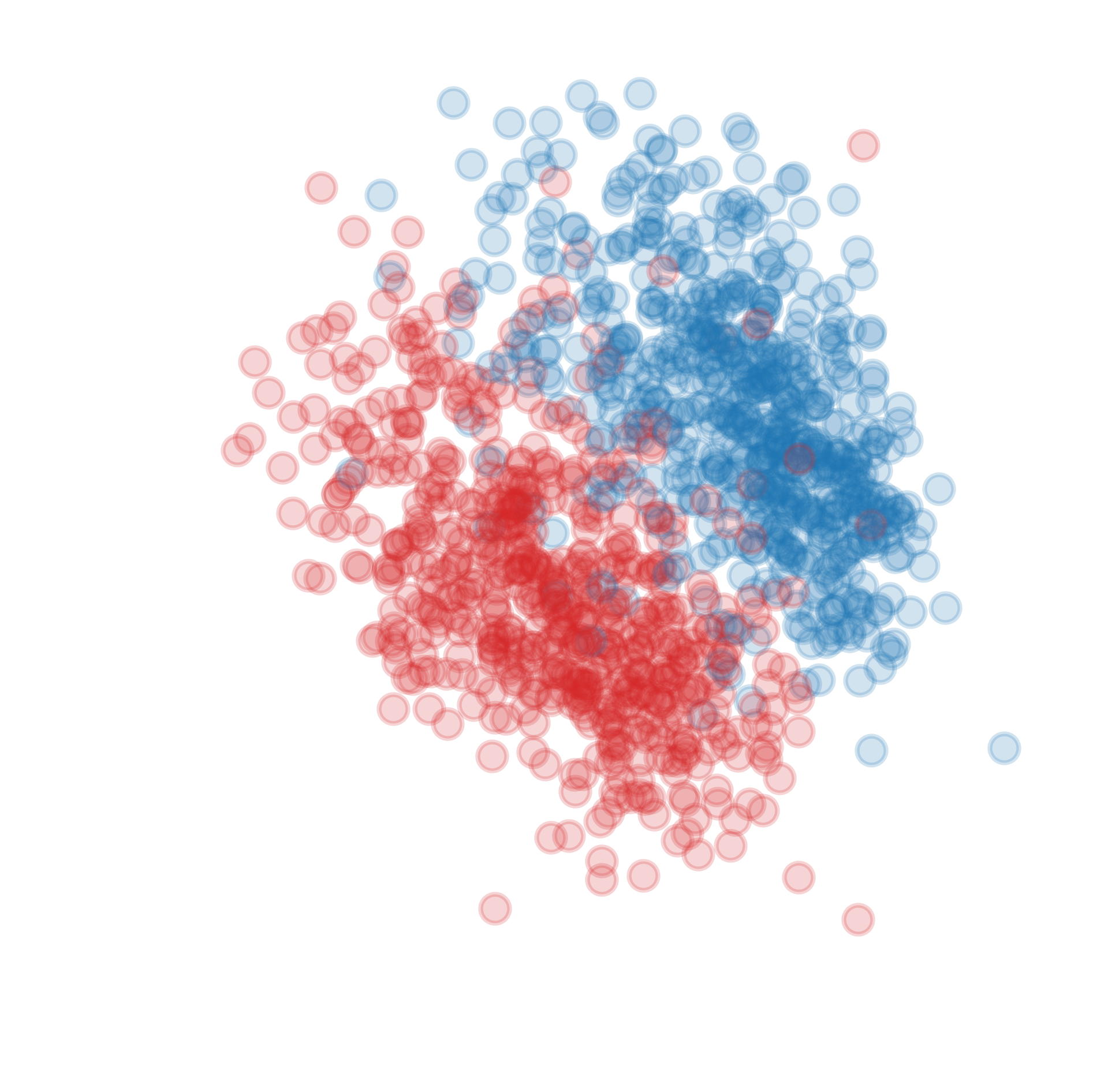}} 
            \label{fig:srs-xstar-2013}
        \end{subfigure}
                \begin{subfigure}[\small $\mathcal{X}$ (2013)]{ 
            \includegraphics[width=0.11\textwidth]{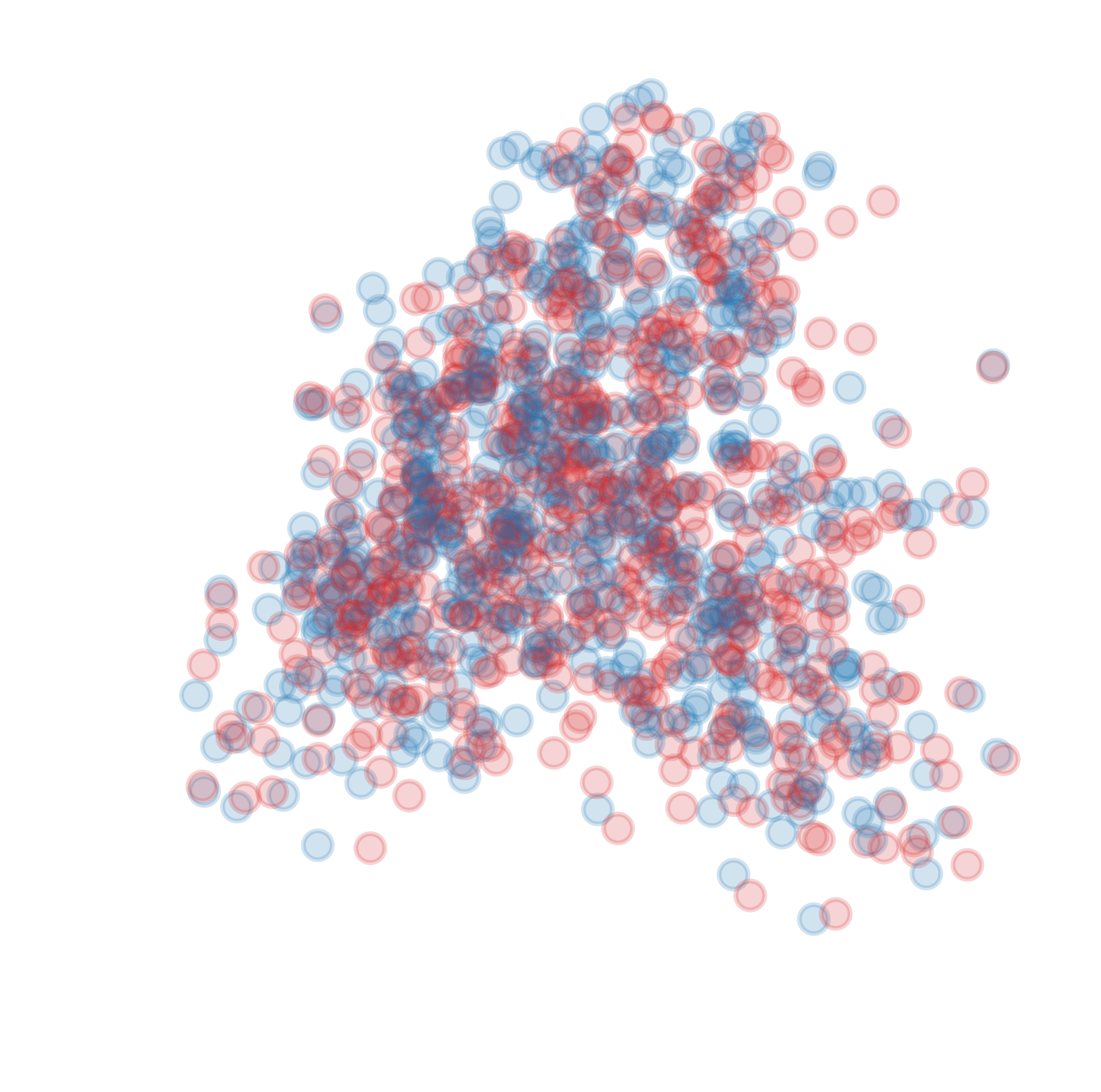}} 
            \label{fig:srs-x-2013}
        \end{subfigure} 
        \begin{subfigure}[\small $\mathcal{X}^\bot_*$ (2013)]{   
            \includegraphics[width=0.11\textwidth]{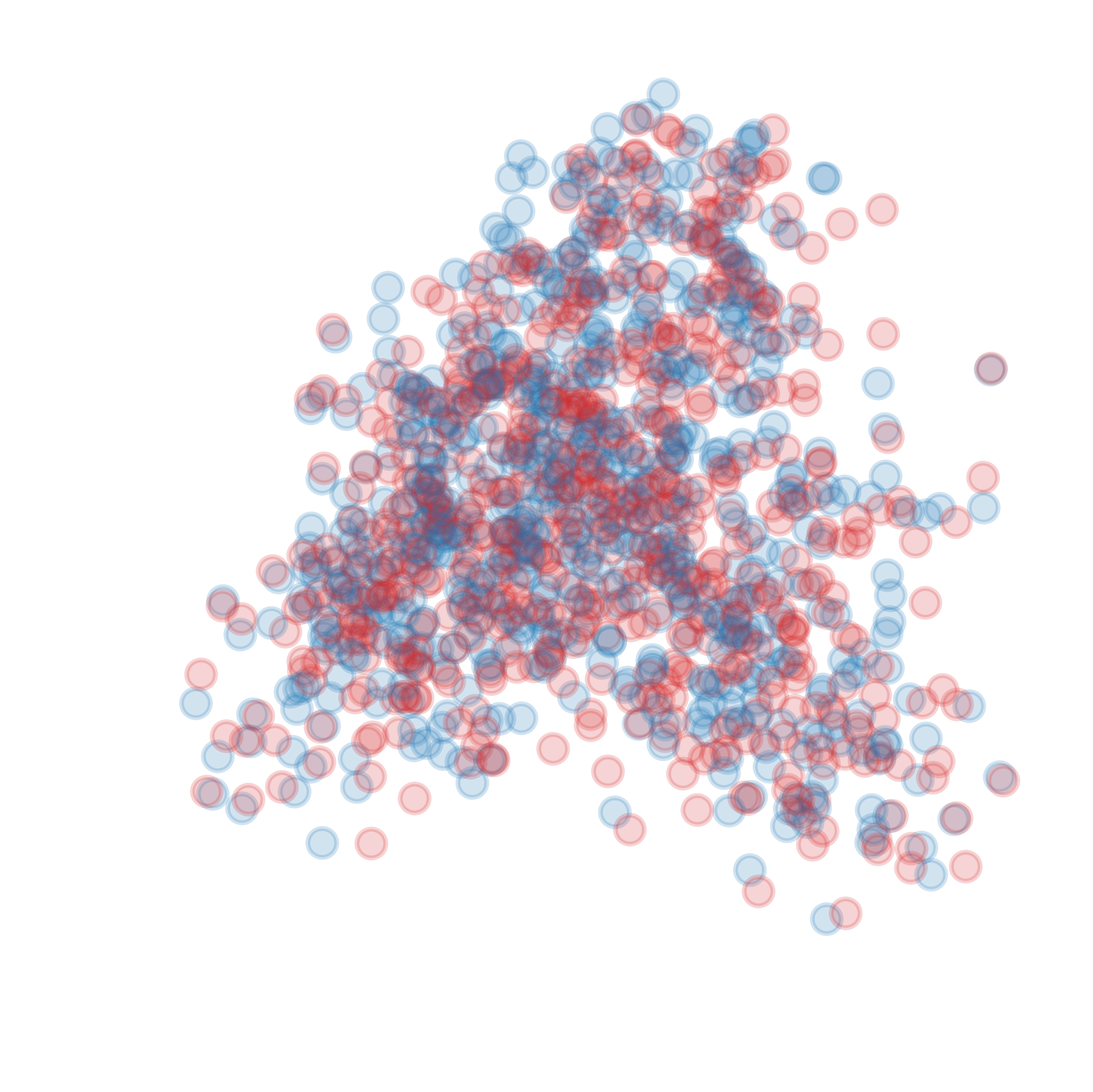}}
            \label{fig:srs-xorth-2013}
        \end{subfigure}      
        \begin{subfigure}[\small $\mathcal{X}_*$ (2014)]{  
            \includegraphics[width=0.11\textwidth]{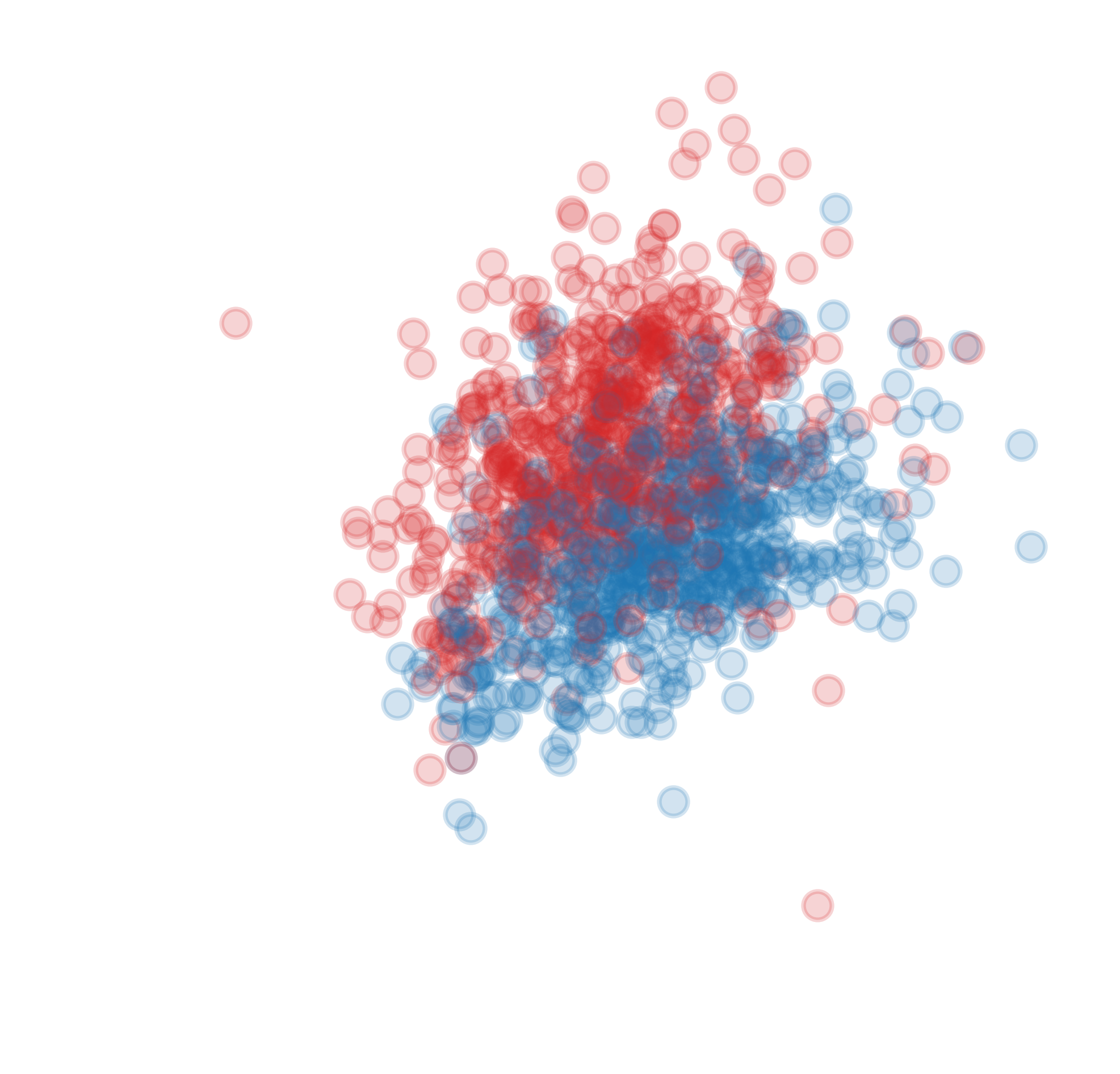}}   
            \label{fig:srs-xstar-2014}
        \end{subfigure}
 \begin{subfigure}[\small $\mathcal{X}$ (2014)]{
            \includegraphics[width=0.11\textwidth]{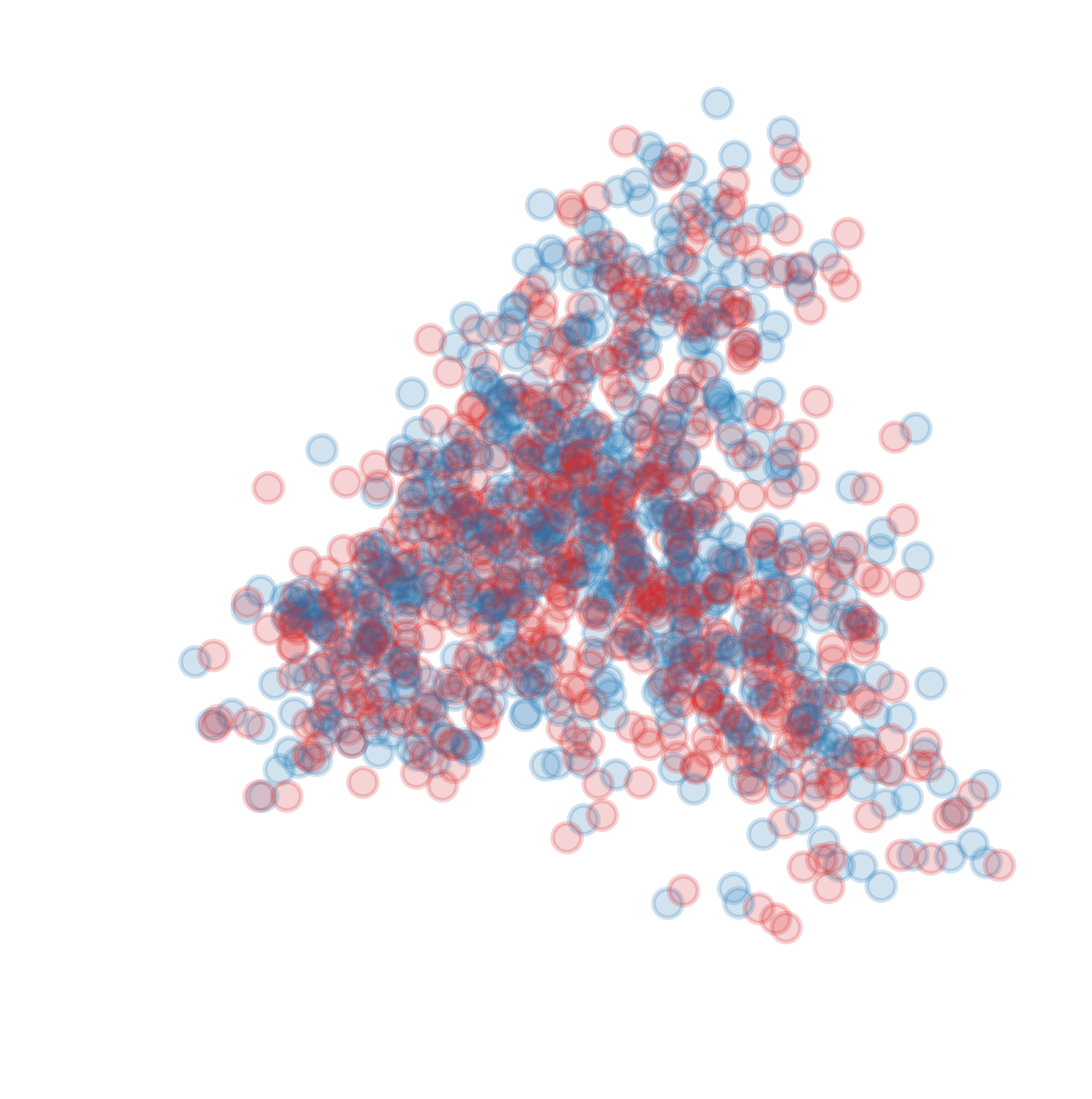}}   
            \label{fig:srs-x-2014}
        \end{subfigure}   
        \begin{subfigure}[\small $\mathcal{X}^\bot_*$ (2014)]{   
            \includegraphics[width=0.11\textwidth]{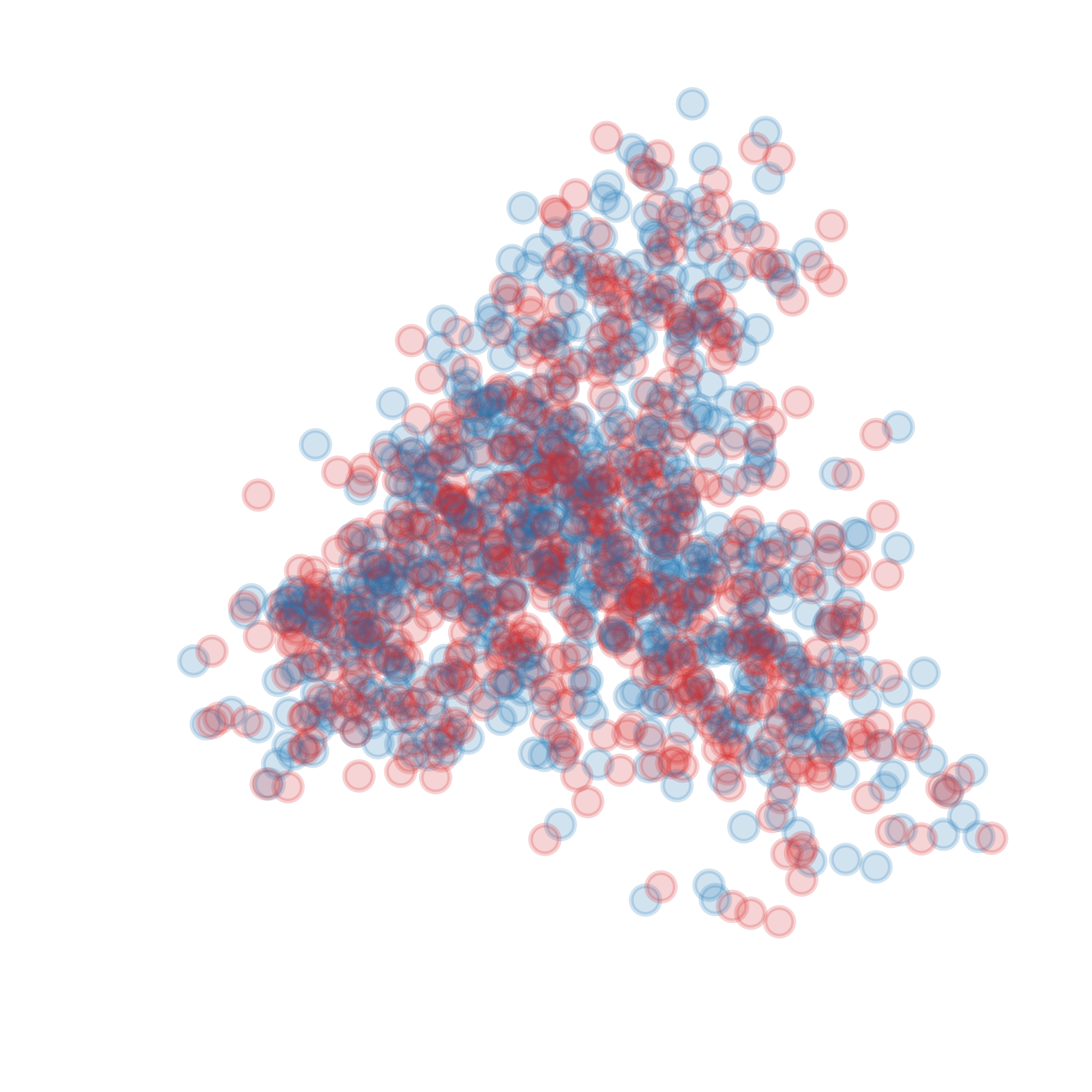}}   
            \label{fig:srs-xorth-2014}
        \end{subfigure}
        \begin{subfigure}[\small $\mathcal{X}_*$ (2015)]{   
            \includegraphics[width=0.11\textwidth]{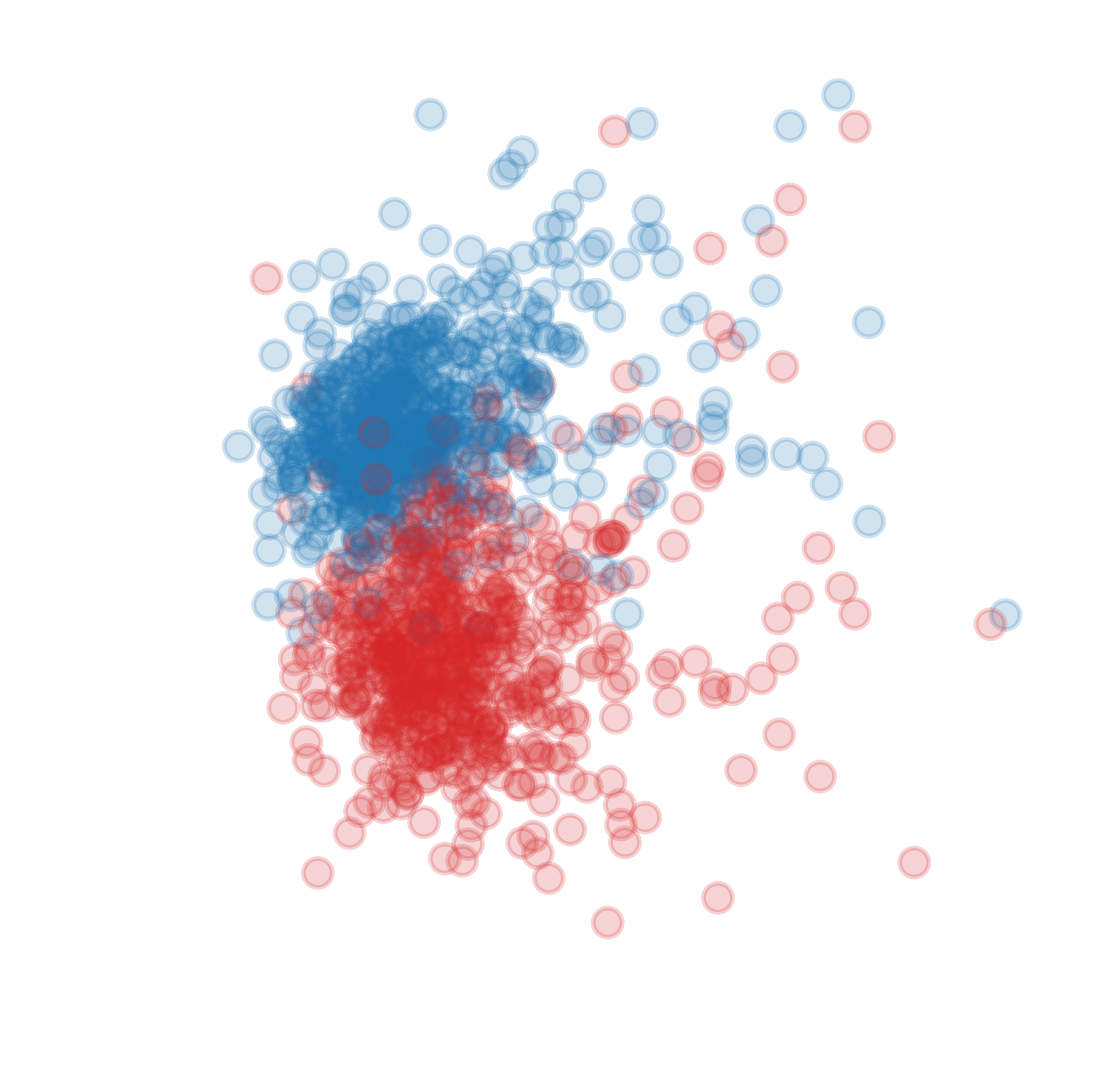}}   
            \label{fig:srs-xstar-2015}
        \end{subfigure}
        \begin{subfigure}[\small $\mathcal{X}$ (2015)]{  
            \includegraphics[width=0.11\textwidth]{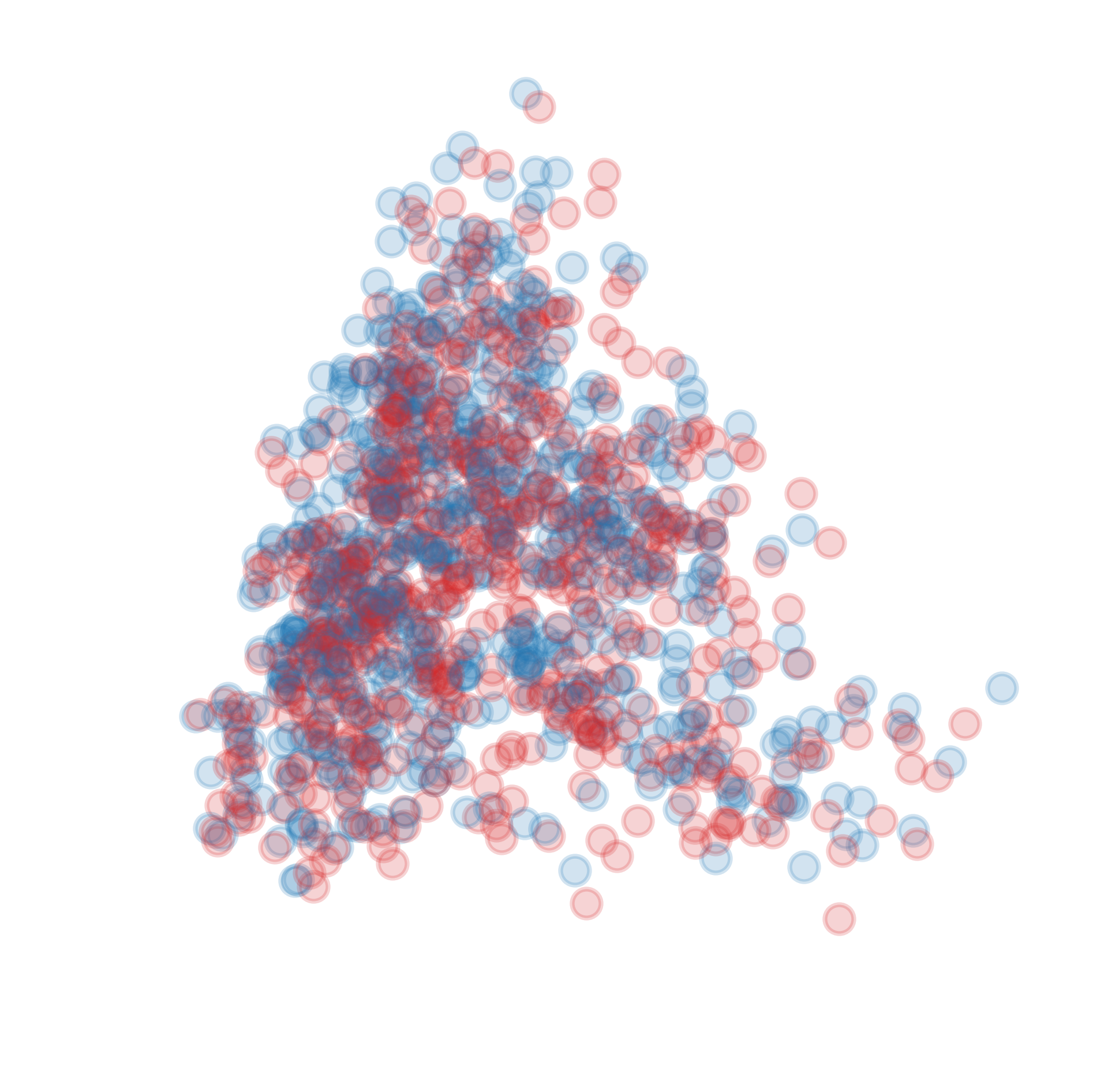}} 
            \label{fig:srs-x-2015}
        \end{subfigure}    
        \begin{subfigure}[\small $\mathcal{X}^\bot_*$ (2015)]{  
            \includegraphics[width=0.11\textwidth]{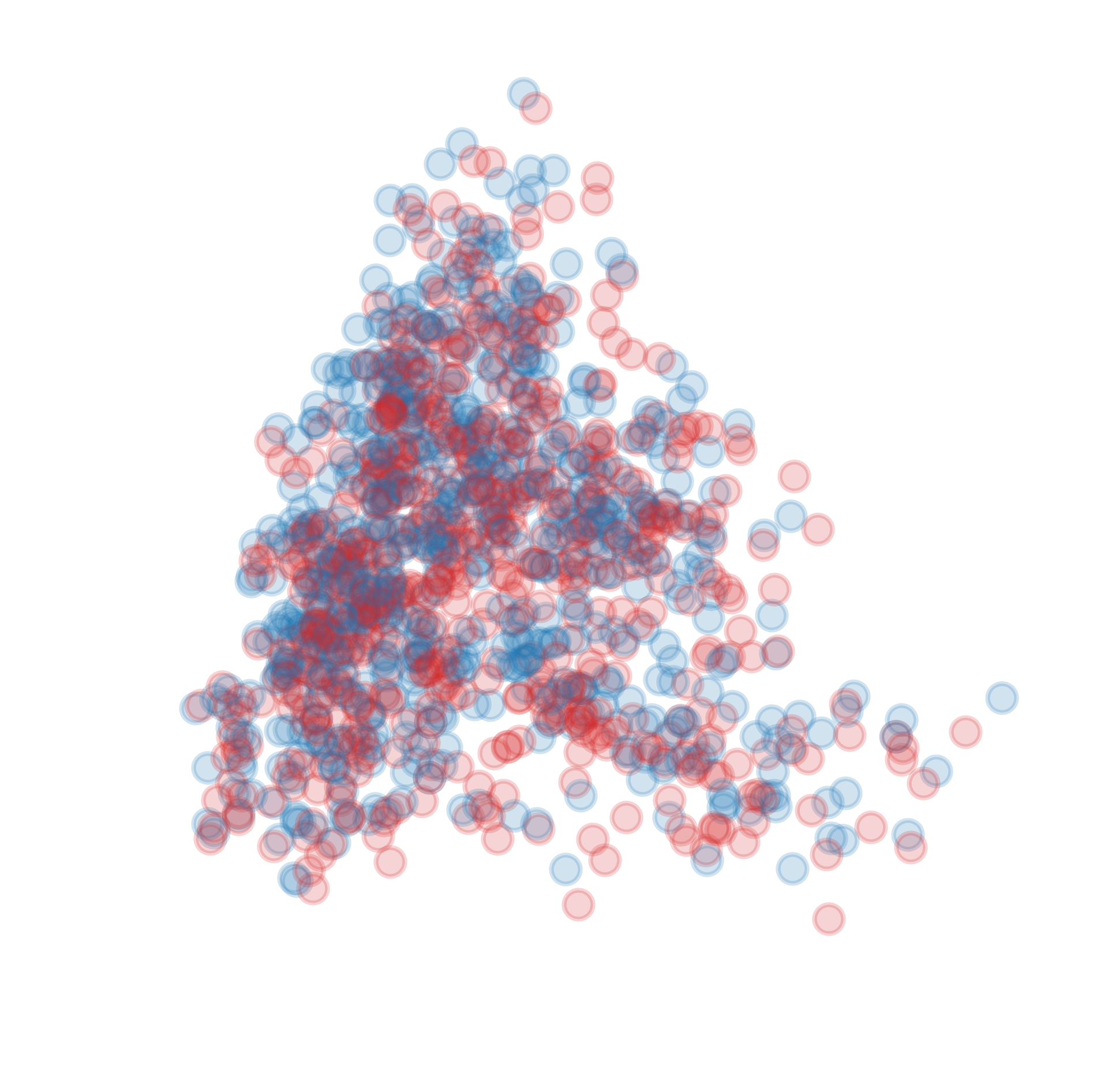}}    
            \label{fig:srs-xorth-2015}
        \end{subfigure}
        \begin{subfigure}[\small $\mathcal{X}_*$ (2016)]{   
            \includegraphics[width=0.11\textwidth]{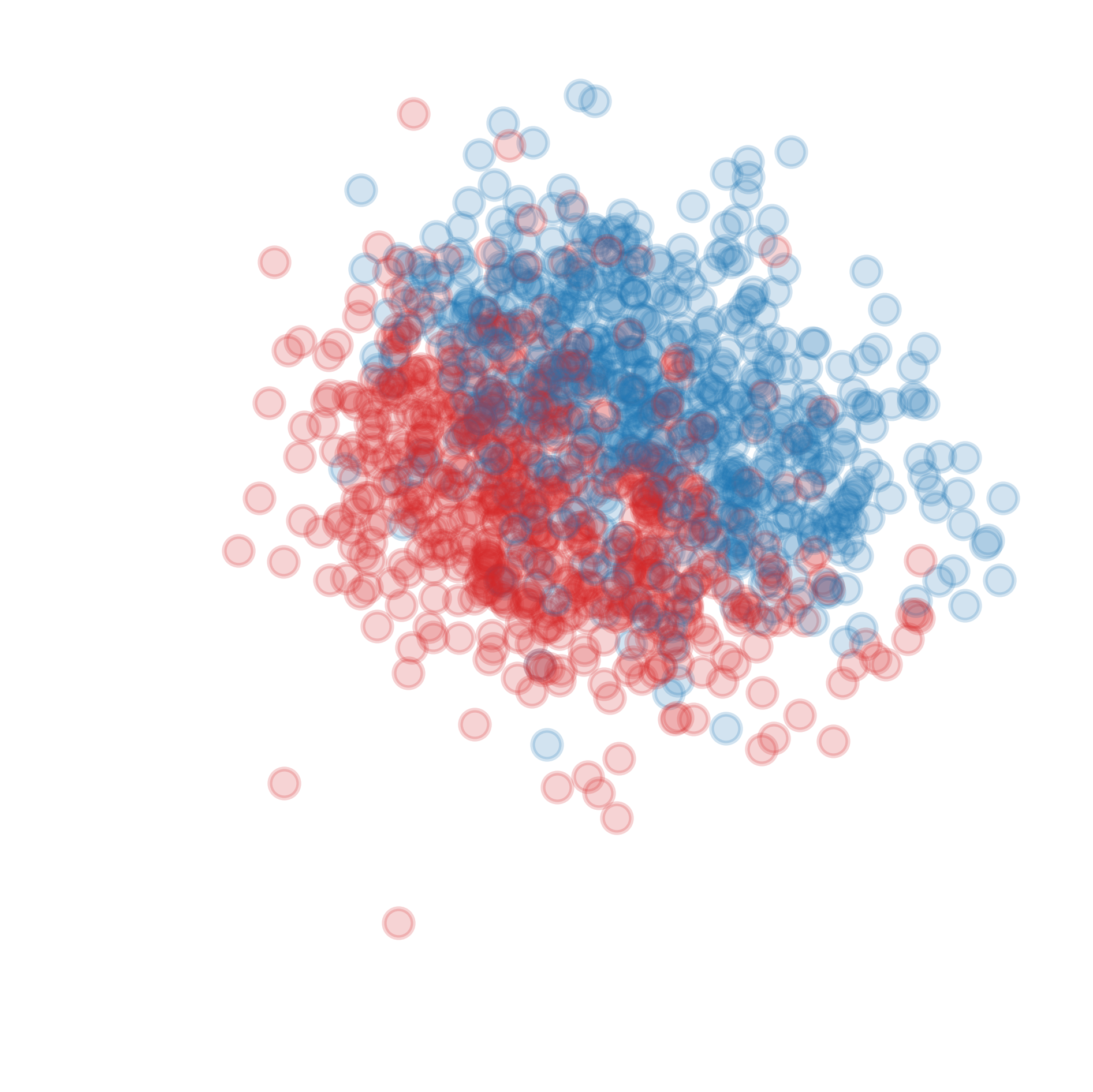}  }  
            \label{fig:srs-xstar-2016}
        \end{subfigure}
        \begin{subfigure}[\small $\mathcal{X}$ (2016)]{
            \includegraphics[width=0.11\textwidth]{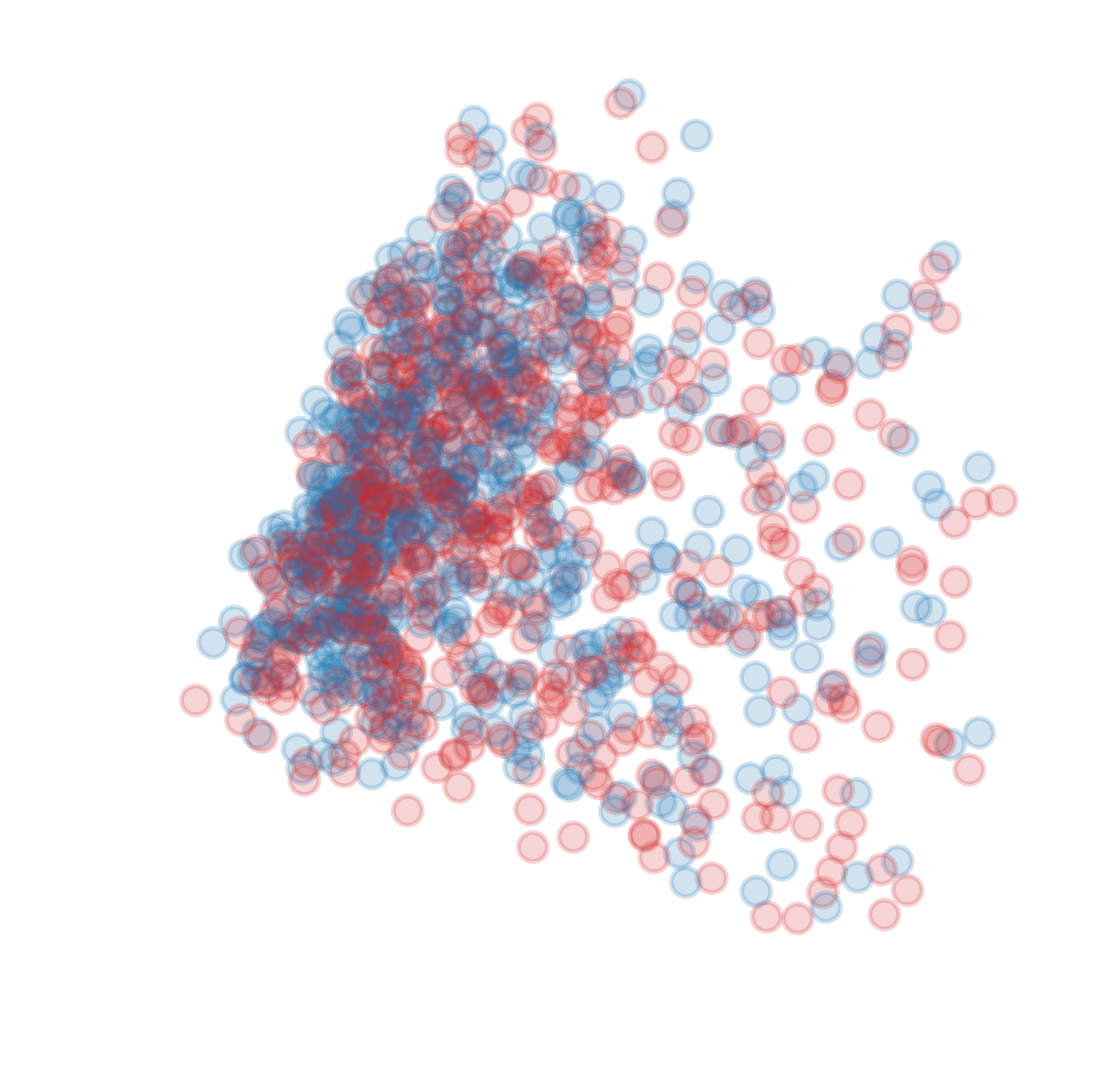} }
            \label{fig:srs-x-2016}
        \end{subfigure}    
        \begin{subfigure}[\small $\mathcal{X}^\bot_*$ (2016)]{ 
            \includegraphics[width=0.11\textwidth]{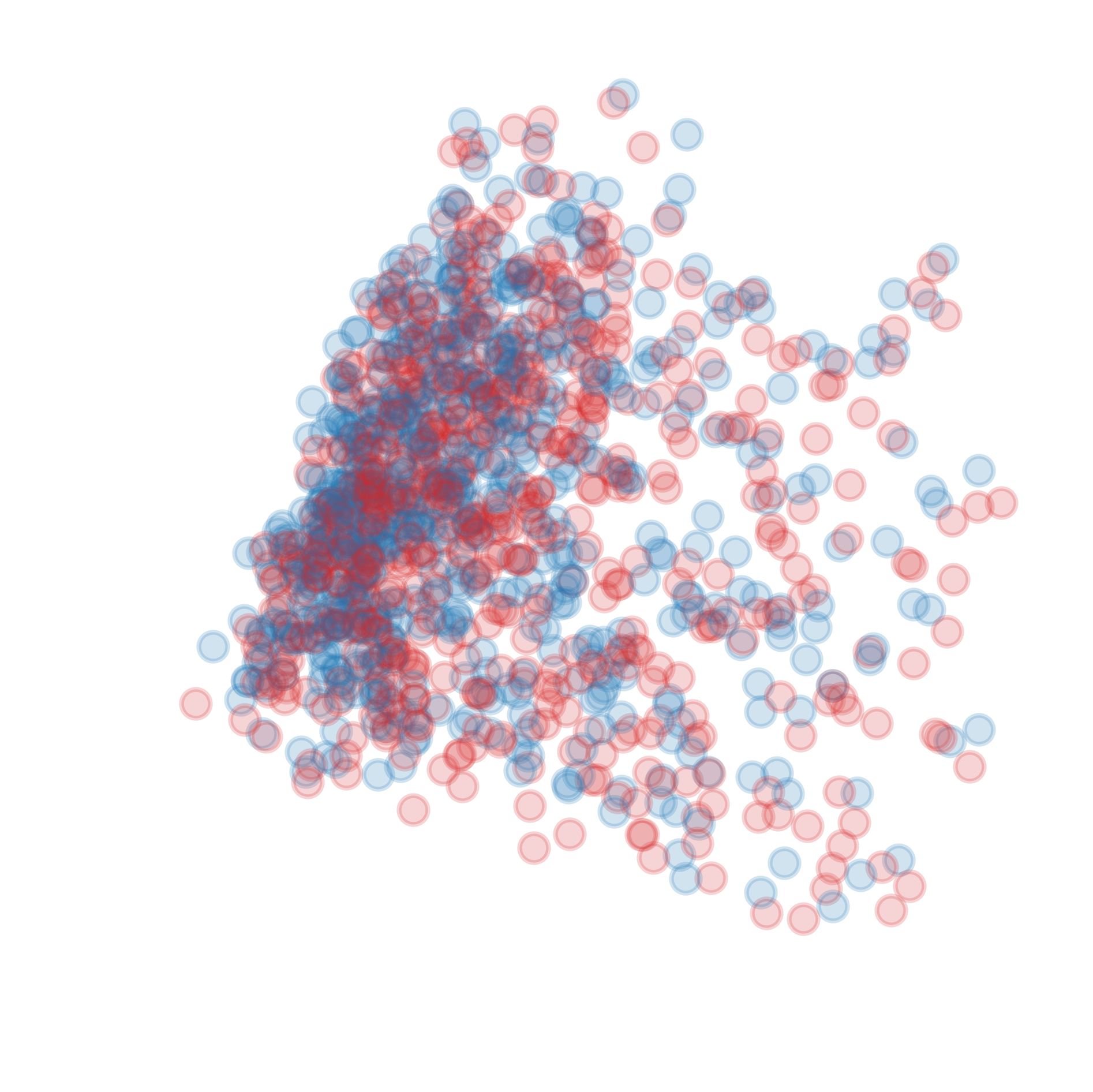}  }
            \label{fig:srs-xorth-2016}
        \end{subfigure}    
        \begin{subfigure}[\small $\mathcal{X}_*$ (2017)]{ 
            \includegraphics[width=0.11\textwidth]{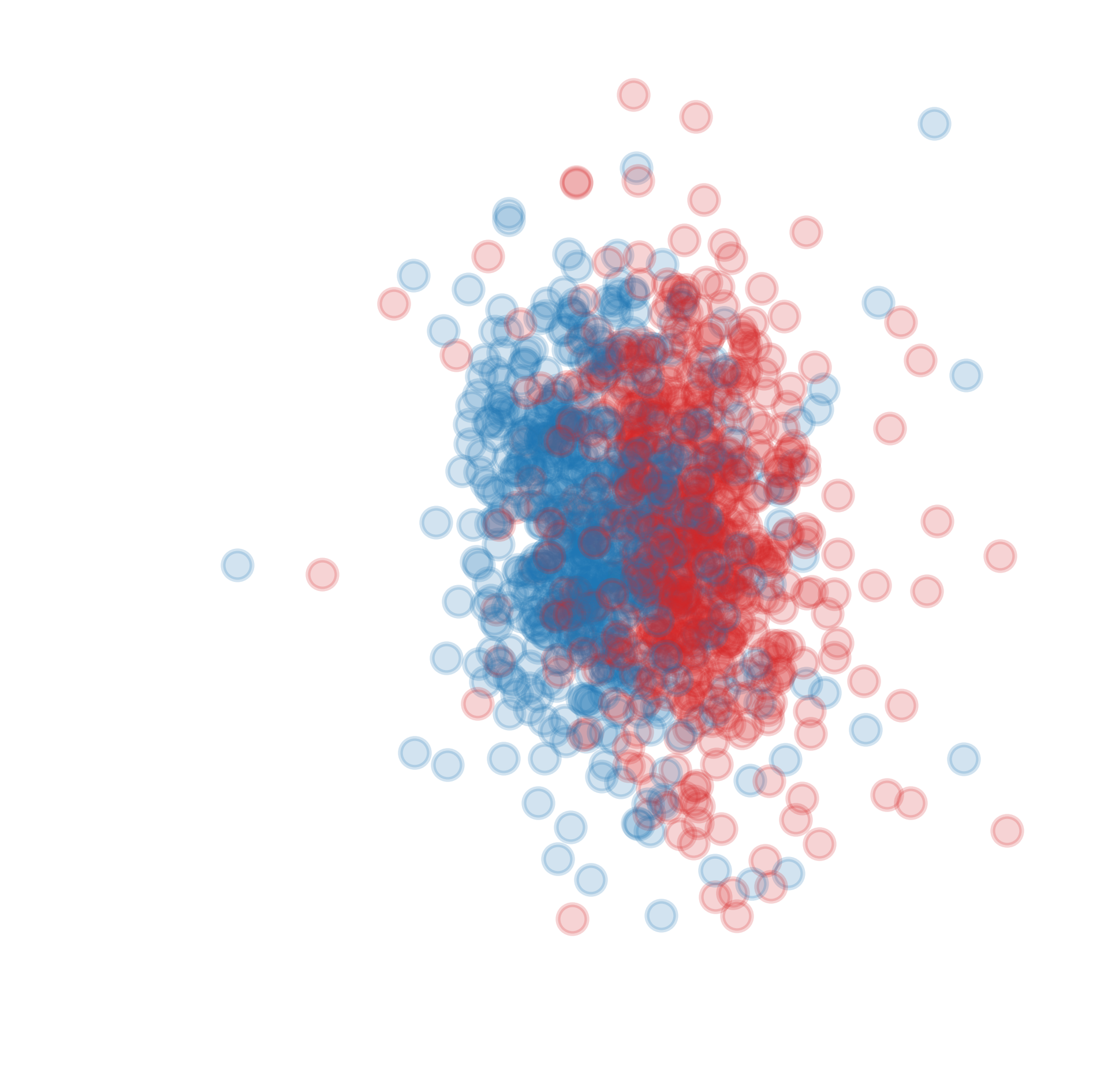}}  
            \label{fig:srs-xstar-2017}
        \end{subfigure}
        \begin{subfigure}[\small $\mathcal{X}$ (2017)]{ 
            \includegraphics[width=0.11\textwidth]{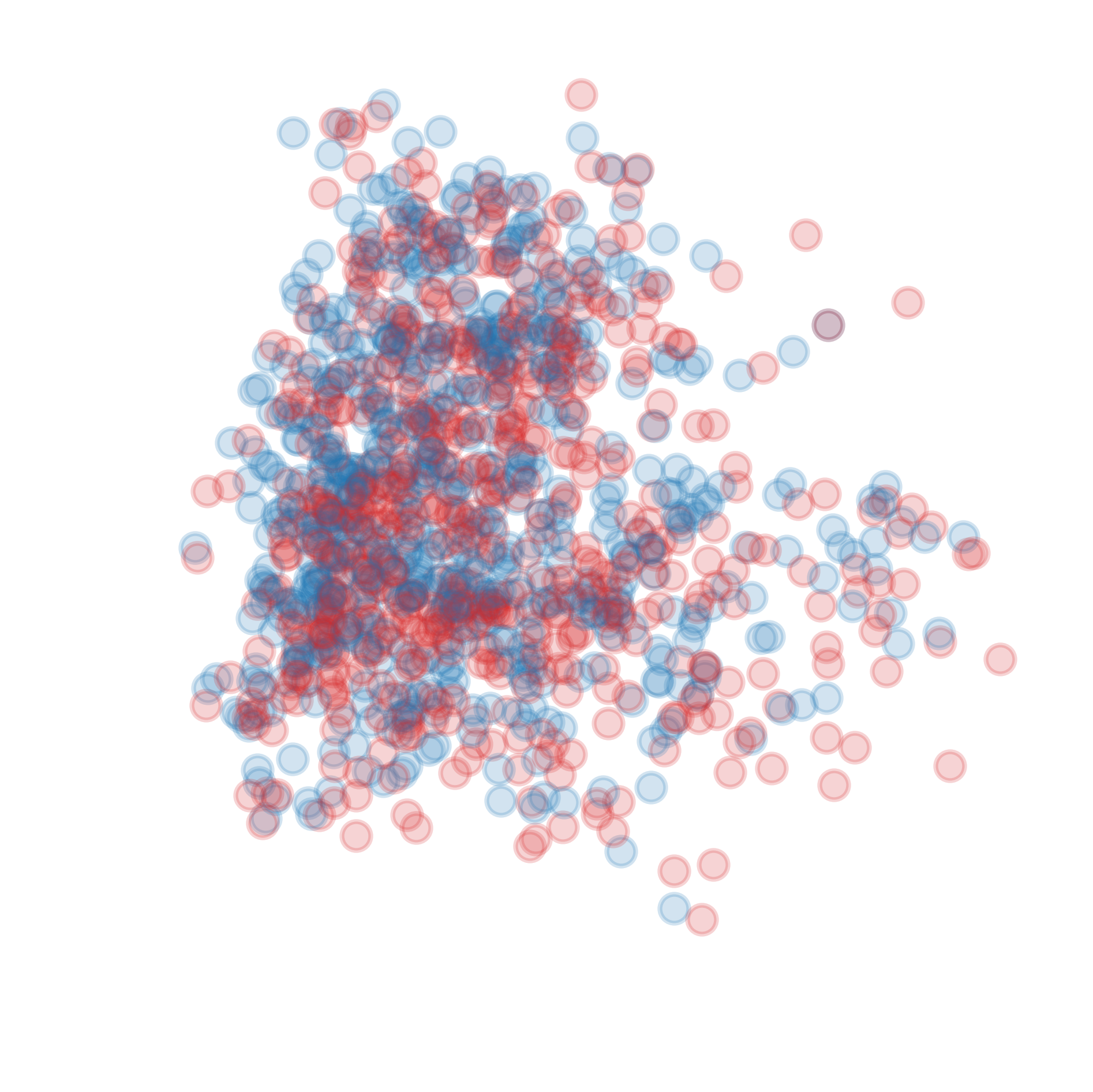}}   
            \label{fig:srs-x-2017}
        \end{subfigure}    
        \begin{subfigure}[\small $\mathcal{X}^\bot_*$ (2017)]{  
            \includegraphics[width=0.11\textwidth]{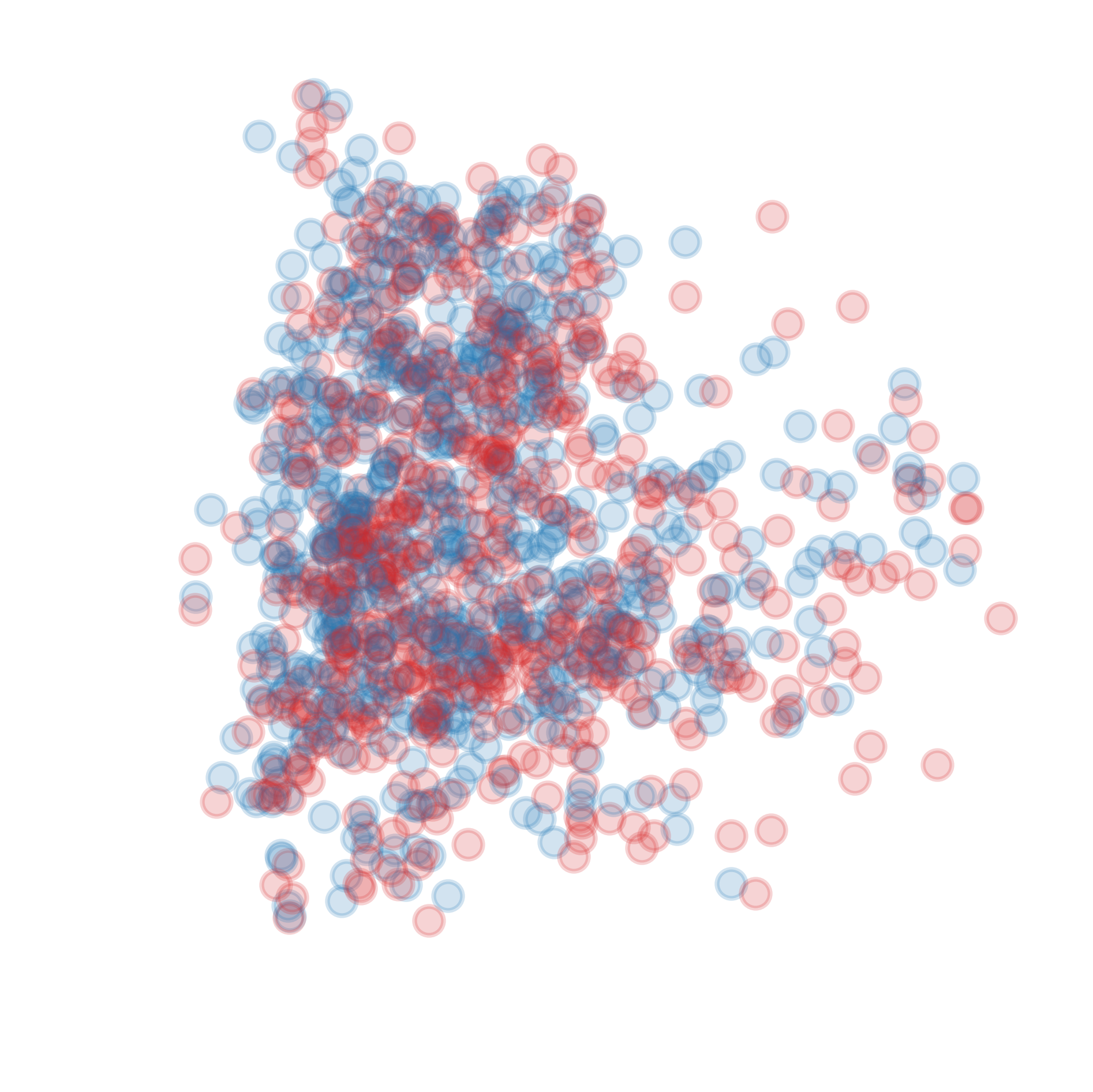}}    
            \label{fig:srs-xorth-2017}
        \end{subfigure}          
        \begin{subfigure}[\small $\mathcal{X}_*$ (2018)]{  
            \includegraphics[width=0.11\textwidth]{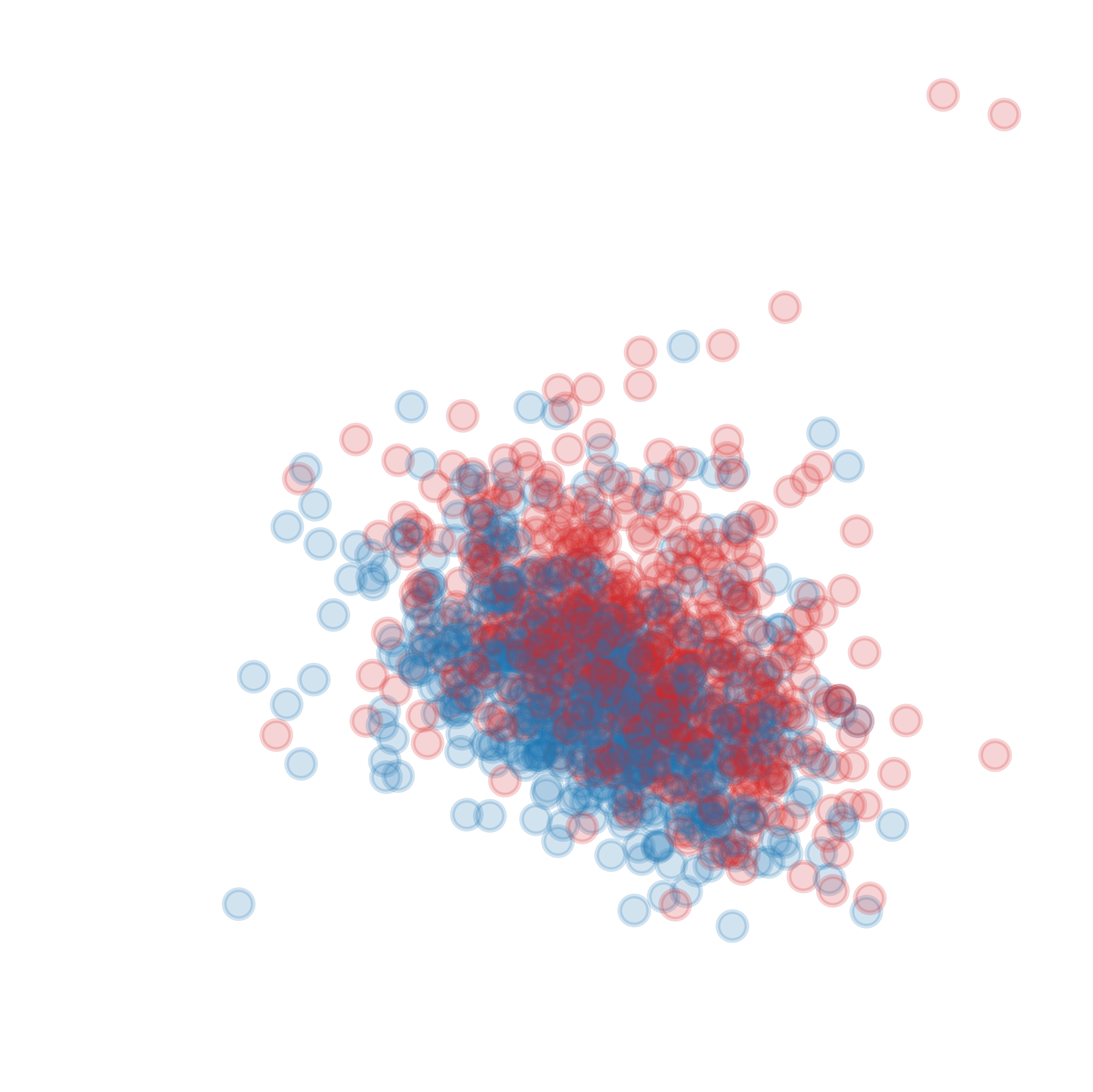}}   
            \label{fig:srs-xstar-2018}
        \end{subfigure}
        \begin{subfigure}[\small $\mathcal{X}$ (2018)]{ 
            \includegraphics[width=0.11\textwidth]{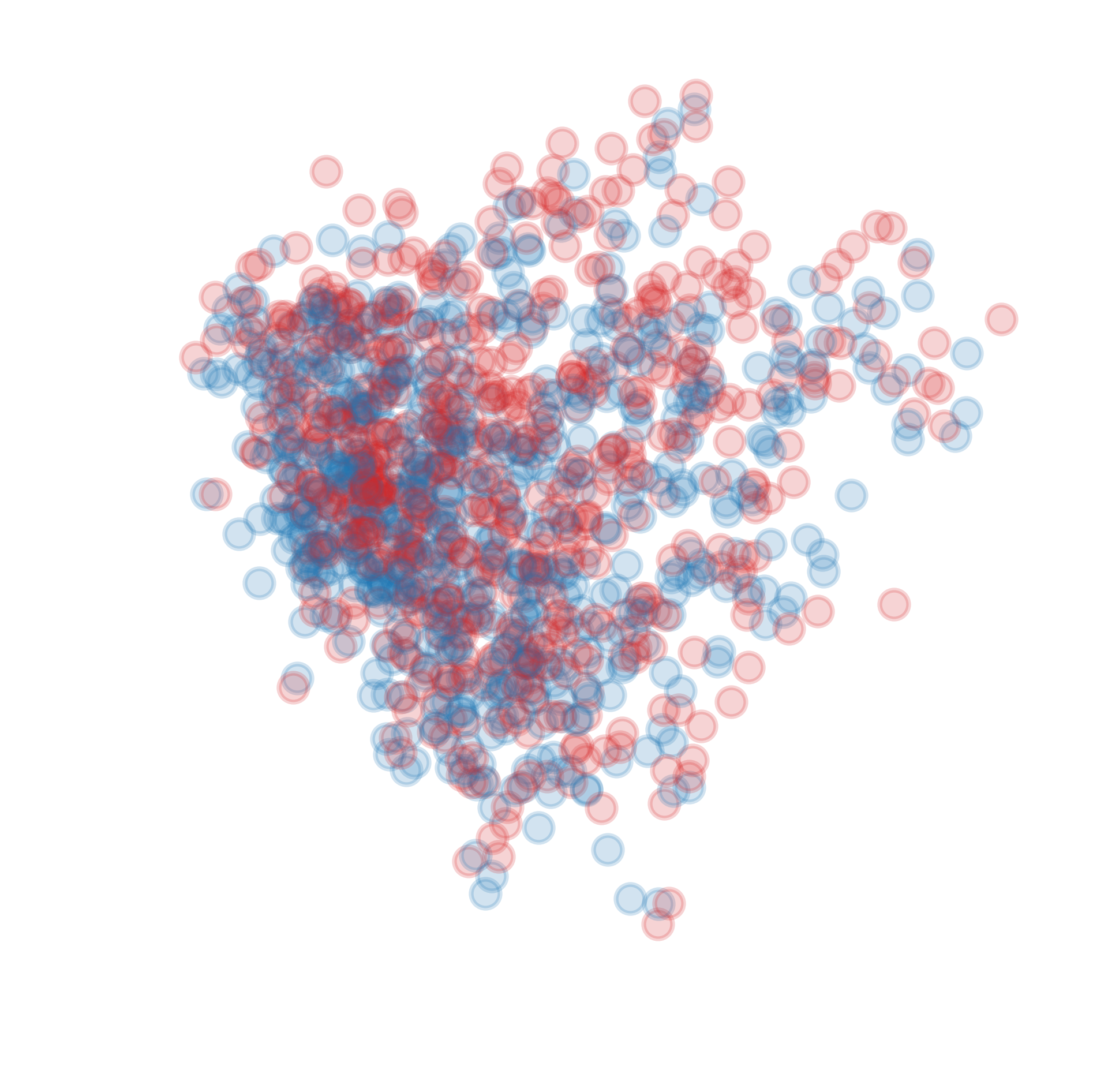}}   
            \label{fig:srs-x-2018}
        \end{subfigure}    
        \begin{subfigure}[\small $\mathcal{X}^\bot_*$ (2018)]{ 
            \includegraphics[width=0.11\textwidth]{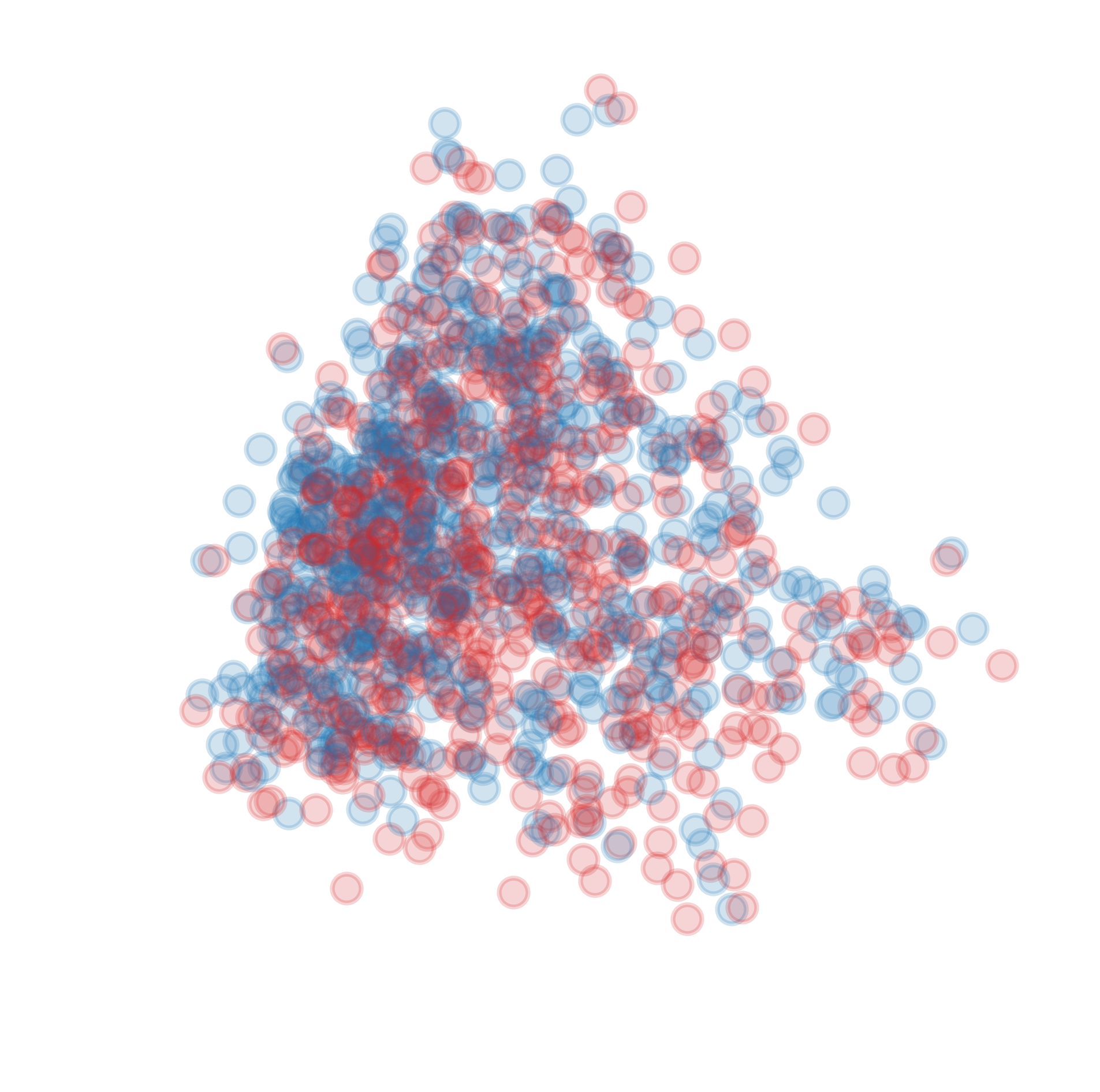}}  
            \label{fig:srs-xorth-2018}
        \end{subfigure}              
        \begin{subfigure}[\small $\mathcal{X}_*$ (2019)]{   
            \includegraphics[width=0.11\textwidth]{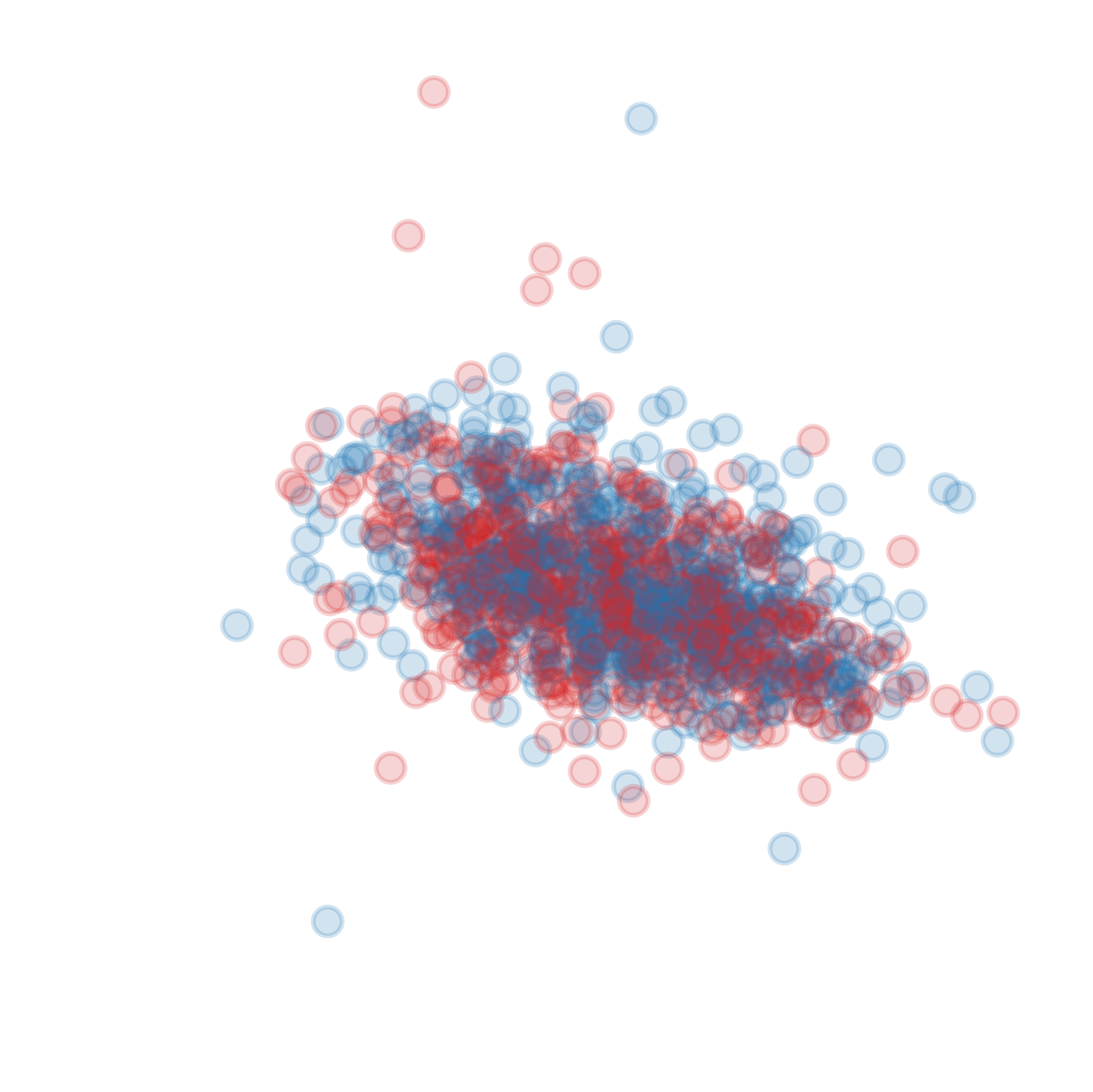}}    
            \label{fig:srs-xstar-2019}
        \end{subfigure}
        \begin{subfigure}[\small $\mathcal{X}$ (2019)]{  
            \includegraphics[width=0.11\textwidth]{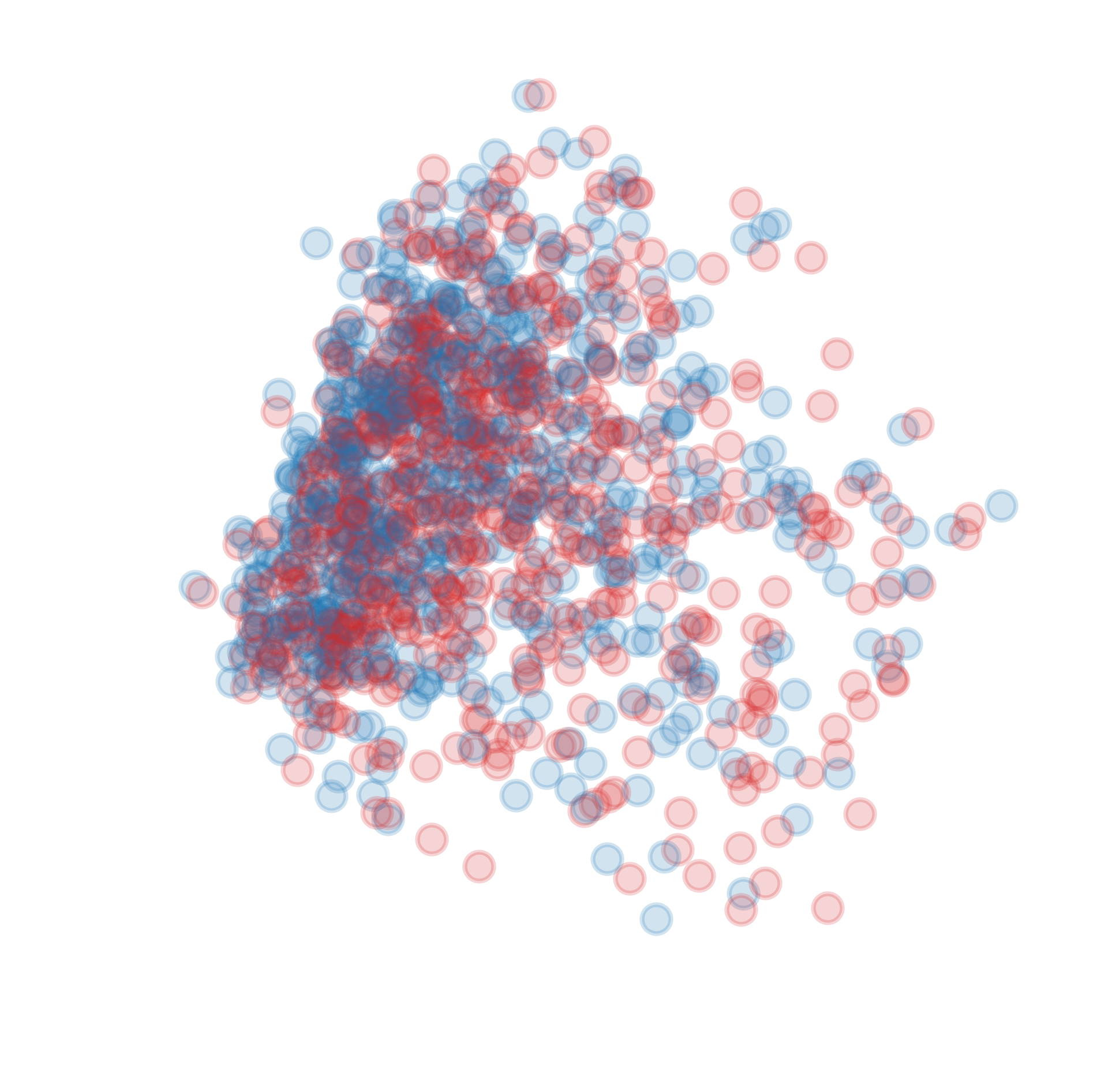}}   
            \label{fig:srs-x-2019}
        \end{subfigure}    
        \begin{subfigure}[\small $\mathcal{X}^\bot_*$ (2019)]{ 
            \includegraphics[width=0.11\textwidth]{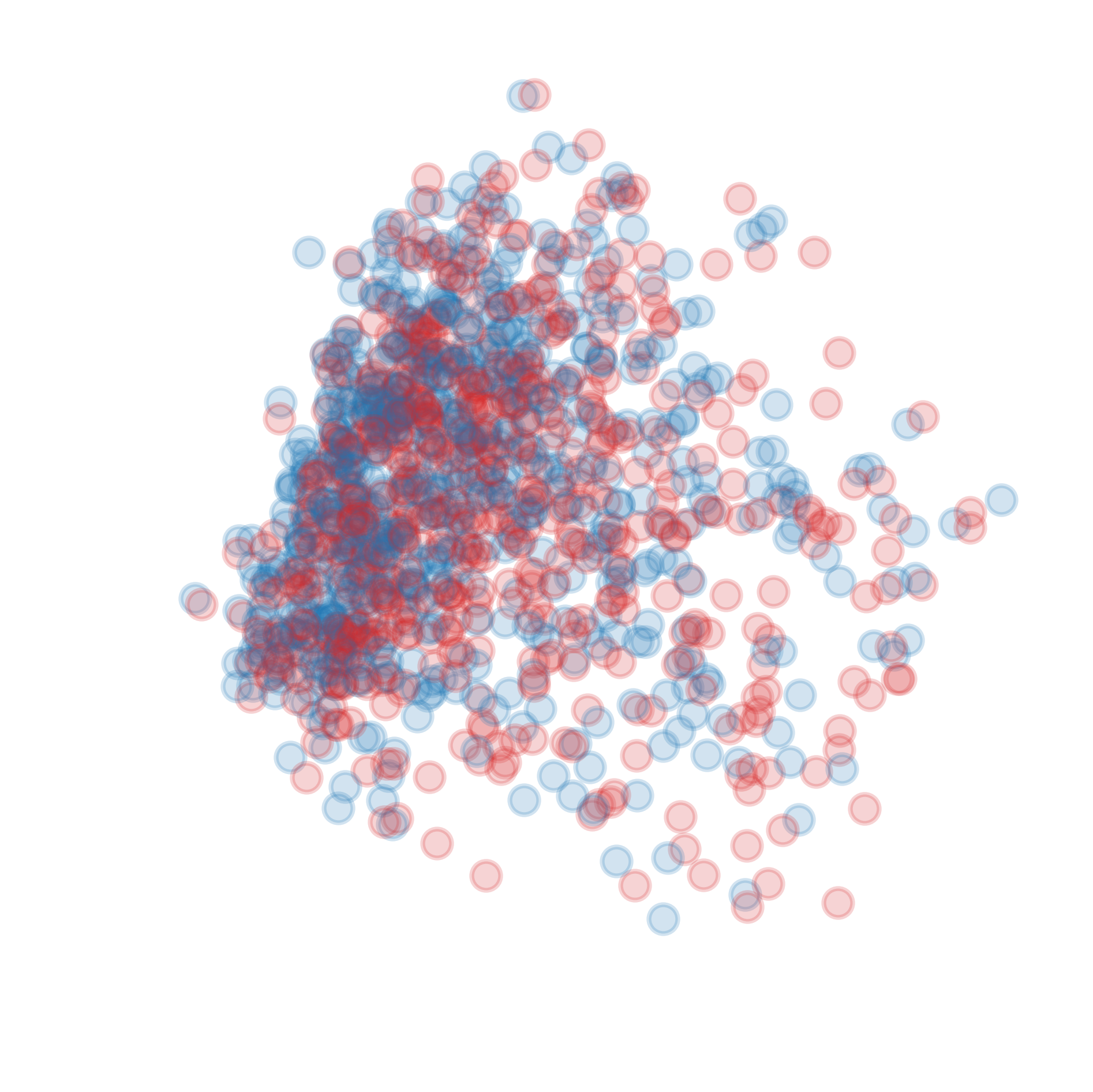}} 
            \label{fig:srs-xorth-2019}
        \end{subfigure}
        \caption[]{$\mathcal{X}$, $\mathcal{X}_*$, and $\mathcal{X}^\bot_*$. $\mathcal{X}_*$ exhibits a clustering
        of embeddings into ideologically left (blue) and right (red). This is not the case for $\mathcal{X}$ and $\mathcal{X}^\bot_*$. The clustering of $\mathcal{X}_*$ into left and right becomes less pronounced after 2016.}
        \label{fig:clustering}
\end{figure}

\begin{table*} [t]
\caption{Performance on ideology prediction (accuracy). The best performance per column (gray) is underlined if it is significantly ($p < .01$) better than the second-best performance as
shown by a McNemar's test \citep{McNemar.1947}.}  \label{tab:performance-logreg}
\centering
\resizebox{\linewidth}{!}{%
\begin{tabular}{@{}lrrrrrrrrrrrrrrrr@{}}
\toprule
{} & \multicolumn{8}{c}{Dev} & \multicolumn{8}{c}{ Test } \\
\cmidrule(lr){2-9}
\cmidrule(l){10-17}
Space & 2013 & 2014 & 2015 & 2016 & 2017 & 2018 & 2019 & $\mu\pm\sigma$ & 
2013 & 2014 & 2015 & 2016 & 2017 & 2018 & 2019 & $\mu\pm\sigma$\\
\midrule
$\mathcal{X}_*$ & \best{.938} & \best{{\underline{.954}}} & \best{{\underline{.942}}} & \best{{\underline{.879}}} & .754 & .646 & .504 & .802$\pm$.161 & \best{{\underline{.946}}} & \best{{\underline{.954}}} & \best{{\underline{.925}}} & \best{{\underline{.904}}} & \best{.783} & .625 & .621 & .823$\pm$.137\\ 
$\mathcal{X}$ & .892 & .887 & .858 & .767 & \best{.804} & \best{{\underline{.796}}} & .787 & .827$\pm$.047 & .887 & .842 & .850 & .792 & .779 & \best{{\underline{.817}}} & .838 & .829$\pm$.034\\ 
$\mathcal{X}^\bot_*$ & .754 & .812 & .692 & .688 & .779 & .783 & \best{.792} & .757$\pm$.046 & .700 & .750 & .754 & .717 & .746 & .775 & \best{.842} & .755$\pm$.042\\ 
\bottomrule
\end{tabular}}
\end{table*}

To probe the semantic information encoded by $\mathcal{X}_*$, we 
draw upon AntSyn \citep{Nguyen.2016b, Nguyen.2017}, a dataset of antonym (and synonym) pairs containing POS information,
which we use to create semantic axes.\footnote{We tried other datasets \citep{Shwartz.2017, An.2018} but found AntSyn to work best for our use case.} More specifically, for each pair of antonyms $a = (p, q)$ (e.g., \textit{small/big}), 
we compute year-wise average contextualized embeddings $\mathbf{x}^{(p)}$ and $\mathbf{x}^{(q)}$, using pretrained (base, uncased) BERT \citep{Devlin.2019} and pooling across subreddits. We discard antonym pairs unless both $p$ and $q$ occur at least 100 times in each year. AntSyn often contains several competing antonym pairs, some of which can be highly context-specific (e.g., \textit{conventional/unconventional} and \textit{conventional/nuclear}). As a simple method to determine the 
most general antonym pair in such cases (e.g., \textit{conventional/unconventional}), we only keep an antonym pair $a$ if $p$ and $q$ are nearest neighbors of each other, resulting in a final set of 972 antonym pairs. We then map $\mathbf{x}^{(p)}$ (and analogously $\mathbf{x}^{(q)}$) into 
the ideological subspace $\mathcal{X}_*$ by computing
\begin{equation}
 \mathbf{x}^{(p)}_* =   \mathbf{P}^\top \mathbf{R}^\top \mathbf{x}^{(p)},
\end{equation}
where $\mathbf{R}$ is the learned (year-specific) rotation matrix, and $\mathbf{P} \in \mathbb{R}^{d \times d_*}$ is the projection
matrix resulting from the learned (year-specific) sparsity pattern in $\mathbf{W}^{(0)}$. We can then define as
\begin{equation}
s_a = \| \mathbf{x}^{(p)}_* \|_2 + \| \mathbf{x}^{(q)}_* \|_2
\end{equation}
a score measuring the importance of the semantic axis imposed by $a$ for the semantic information captured by the ideological subspace $\mathcal{X}_*$.\footnote{We tried other formulations and obtained similar results.} Large (small) values of $s_a$ indicate that $\mathcal{X}_*$ captures much (little) of $a$'s semantics.

We first examine quantitatively whether there are systematic patterns regarding the lexical semantics of axes that are well captured by $\mathcal{X}_*$ ($s_a$ large) compared to axes where this is not the case ($s_a$ small). Drawing upon crowdsourced datasets 
of affect \citep{Warriner.2013}, concreteness \citep{Brysbaert.2014}, and morality \citep{Hopp.2021} ratings for words (which 
all provide continuous human evaluation scores), we compute values for 
each $a$ by averaging the ratings of $p$ and $q$. Comparing the 100 pairs with top $s_a$ (strongly encoded by $\mathcal{X}_*$) with the 100 pairs with bottom $s_a$ (weakly encoded by $\mathcal{X}_*$), we find that they have consistently lower concreteness and higher morality values (Table \ref{tab:concreteness}), but there is no clear trend for affect (not shown). These results suggest that the ideologically-driven
cooccurrence variations (i.e., differences in framing) that cause ideological bias 
are related to abstract semantics and moral-like reasoning. While the former is a property of political language in general, the latter can be related to recent insights about the central nature
of moral judgments for political ideology \citep{Haidt.2004,Haidt.2007, Graham.2009,  Graham.2013} and framing \citep{Fulgoni.2016, Mokhberian.2020, Mendelsohn.2021}.

Furthermore, we qualitatively inspect pairs with highest and lowest values of $s_a$. Focusing on adjectives (Table~\ref{tab:axes}), 
we find that many of them express either general or specifically political and moral evaluative 
semantics such as \textit{useful/useless}, \textit{biased/impartial}, and \textit{immoral/moral}.
This is  in line with our quantitative analysis, and it is also confirmed by the nominal and verbal axes 
(Appendix \ref{ap:axes}),
with pairs such as \textit{ability/inability}, \textit{patriot/traitor}, and \textit{barbarism/culture}
or \textit{agree/disagree}, \textit{overpay/underpay}, and \textit{dehumanize/humanize}. We also notice that pairs with the 
lowest ranks (e.g., \textit{north/south}, \textit{husband/wife}, and \textit{lock/unlock}) tend not to exhibit this evaluative character.

Thus, $\mathcal{X}_*$ encodes abstract evaluative
 categories, specifically ones that are constitutive of
political reasoning and framing. Furthermore, some pairs explicitly refer to opposing political 
concepts (e.g., \textit{autocratic/democratic}), which begs the question
whether $\mathcal{X}_*$ represents ideology indexically.

\subsection{Indexical Probing of $\mathcal{X}_*$} \label{sec:ideological}

We are interested to see whether the topology of $\mathcal{X}_*$ captures ideology, i.e., 
we want to find systematic patterns induced by the indexical associations of $\mathcal{X}_*$. To do so, we 
examine 
to what extent the embeddings in $\mathcal{X}_*$ exhibit,
for certain ideologies, a cluster structure. 
We focus on the left-right ideological spectrum due to its central importance for US politics \citep{Heywood.2017}.
We draw upon a left-wing (\texttt{socialism}) and a right-wing (\texttt{conservatives})
subreddit that are both among the largest subreddits.\footnote{Results are robust with respect to the 
selection of the subreddits and similar when using other left-wing and right-wing subreddits (e.g., \texttt{communism} and \texttt{Republican}).}
Using all concepts from train and plotting their embeddings for all years by means of PCA (Figure~\ref{fig:clustering}), 
we observe a strong clustering according to ideological groups in $\mathcal{X}_*$ that becomes less pronounced 
after 2016. Furthermore, we do not observe a strong clustering for the PCA plots of $\mathcal{X}$
and $\mathcal{X}^\bot_*$. Notice this ideological split is an \textit{intrinsic property of the embeddings}: our method
does not add any information to the embedding space but finds the subspace
that already contains the ideological bias.

\begin{figure}[t!]
        \captionsetup[subfigure]{aboveskip=-2mm}
        \centering                                           
        \begin{subfigure}[\small $\mathcal{X}_*$ (2019)]{   
            \includegraphics[width=0.11\textwidth]{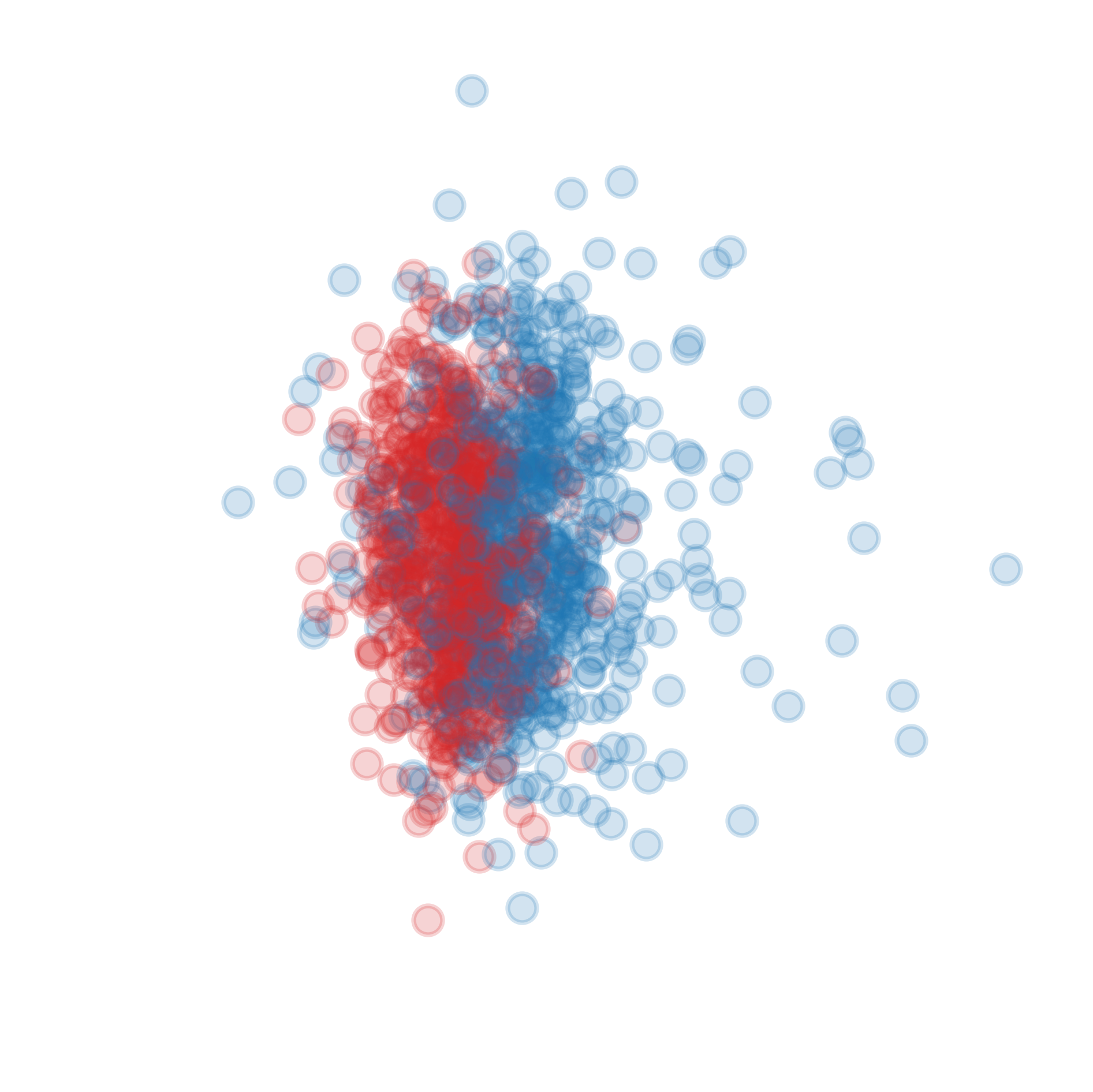}}
            \label{fig:srs-xstar-2019}
        \end{subfigure}
        \begin{subfigure}[\small $\mathcal{X}$ (2019)]{ 
            \includegraphics[width=0.11\textwidth]{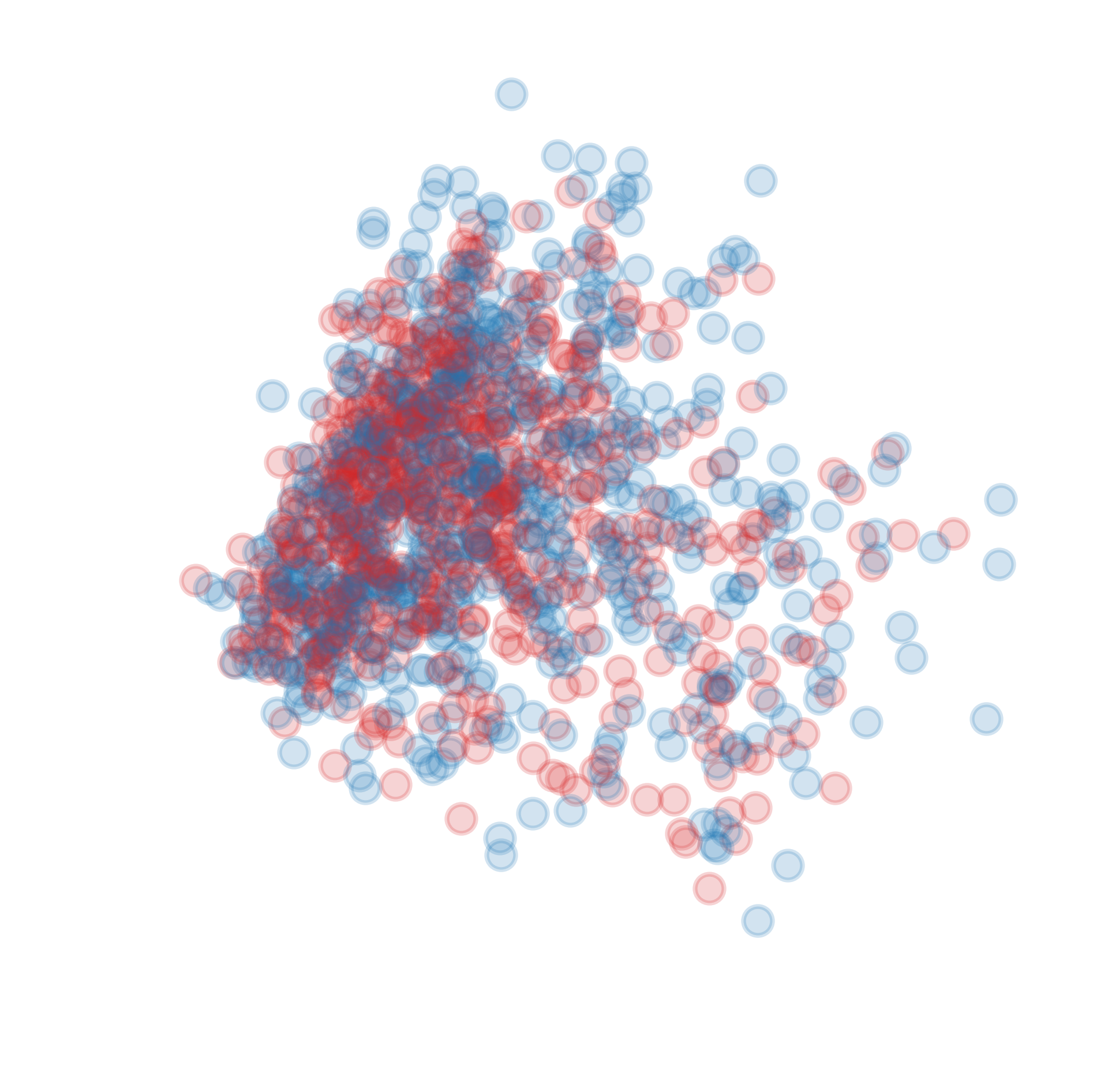}}
            \label{fig:srs-x-2019}
        \end{subfigure}    
        \begin{subfigure}[\small $\mathcal{X}^\bot_*$ (2019)]{
            \includegraphics[width=0.11\textwidth]{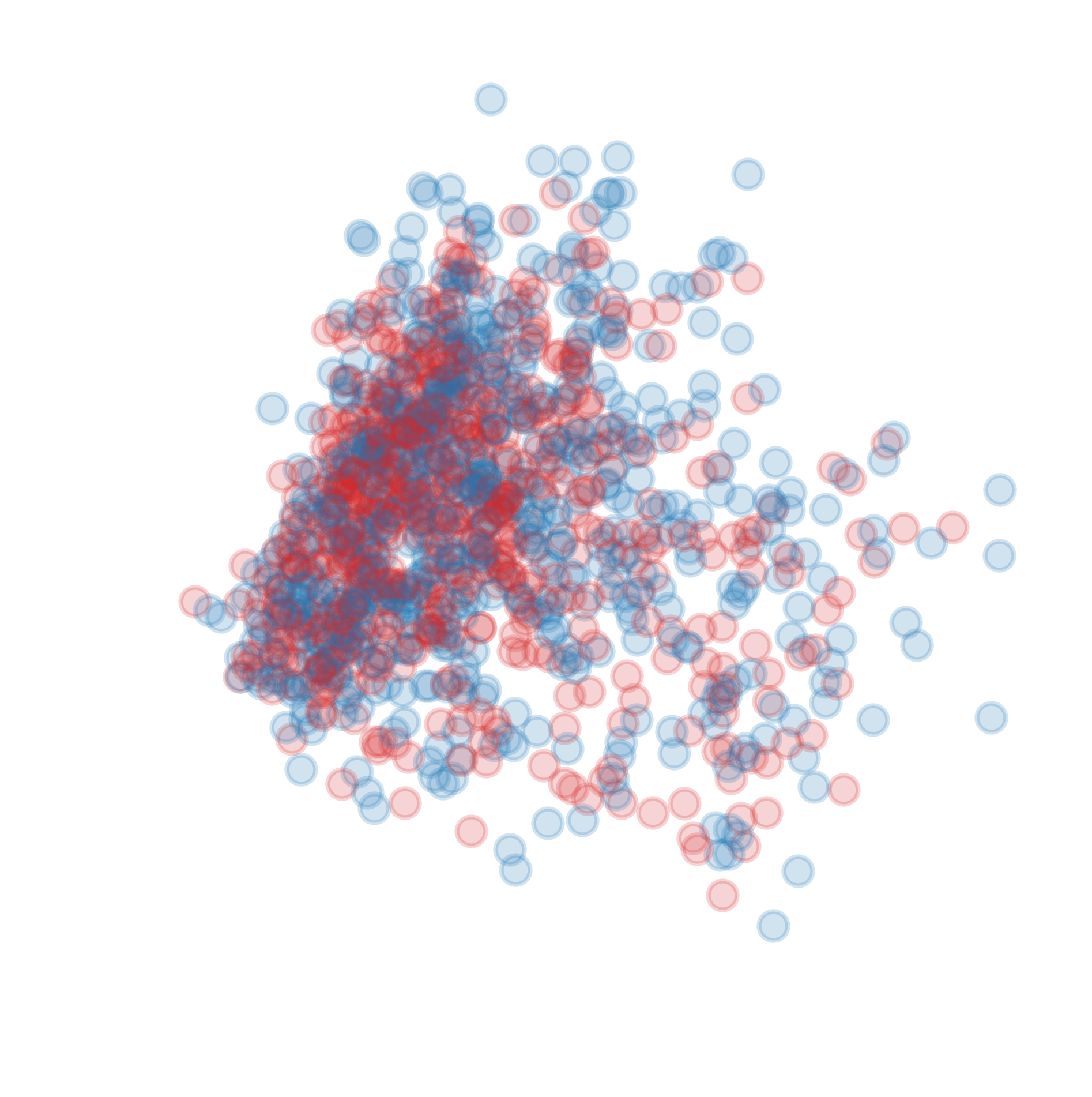}}
            \label{fig:srs-xorth-2019}
        \end{subfigure}
        \caption[]{Embedding space topology of $\mathcal{X}$, $\mathcal{X}_*$, and $\mathcal{X}^\bot_*$. The first two principal components of $\mathcal{X}_*$ exhibit a clustering
        of embeddings into pro-Trump (red) and anti-Trump (blue). This is not the case for $\mathcal{X}$ and $\mathcal{X}^\bot_*$.}
        
        \label{fig:clustering-trump}
\end{figure}

To test this more quantitatively, we split the concept embeddings for both ideologies
into train (60\%), dev (20\%), and test (20\%) and train year-wise logistic regression classifiers to predict the
subreddit ideology (left-wing versus right-wing) from the
concept embeddings. Since we do not need to tune hyperparameters, 
we use the dev sets for additional testing. We compare the performance of the embeddings in $\mathcal{X}_*$
to the ones in $\mathcal{X}$ and $\mathcal{X}^\bot_*$.

We find that (i) the performance of 
the embeddings in $\mathcal{X}_*$ drastically decreases after 2016, and (ii)
that the embeddings in $\mathcal{X}_*$ perform substantially better than the ones in 
 $\mathcal{X}$ and $\mathcal{X}^\bot_*$ from 2013 to 2016, but similarly or worse from 2017 to 2019 (Table \ref{tab:performance-logreg}).
 This confirms the observation (Figure~\ref{fig:clustering}) that the left-right spectrum as reflected by $\mathcal{X}_*$
 becomes less pronounced after 2016. The embeddings in $\mathcal{X}^\bot_*$ generally perform worse than the ones
 in $\mathcal{X}$, indicating that they contain less ideologically relevant information.

 Why does the ideological left-right spectrum as reflected by $\mathcal{X}_*$ become less pronounced over time? 
 It is striking to observe that 
 the decreasing trend seems to be parallel with the inception of the presidency of Donald Trump.
 In fact, when we repeat 
the analysis for representative pro-Trump (\texttt{The\_Donald}) and anti-Trump (\texttt{AntiTrumpAlliance})
subreddits, we see a separation for the years after 2016 (Figure \ref{fig:clustering-trump}).\footnote{
Results are again robust with respect to the 
selection of the subreddits and similar when using other pro-Trump and anti-Trump subreddits (e.g., \texttt{trump} and \texttt{Impeach\_Trump}).} It is well known that Trump has profoundly impacted the political discourse on Reddit \citep{Massachs.2020}.
 At the same time, Trump is notoriously hard to assign a position on the left-right spectrum \citep{Carmines.2016, Barber.2019}.
 Taken together, this suggests that the dominating ideological axis in the Reddit Politosphere after 2016 is not 
 left versus right but rather pro-Trump versus anti-Trump, which is captured 
 by the indexical structure of $\mathcal{X}_*$. This analysis is also supported by the observation that the performance is only decreasing for 
$\mathcal{X}_*$, but not for $\mathcal{X}$ and $\mathcal{X}^\bot_*$.

The results underscore
 that $\mathcal{X}_*$ contains ideological information from $\mathcal{X}$ in distilled form and influenced by the dominant
 axes of antagonism in the data. If these axes
 change, so does the ideological information captured by
 $\mathcal{X}_*$.

\subsection{Case Study: Dispersion in $\mathcal{X}_*$} \label{sec:dispersion}

\begin{figure}[t!]
        \centering
                \includegraphics[width=0.4\textwidth]{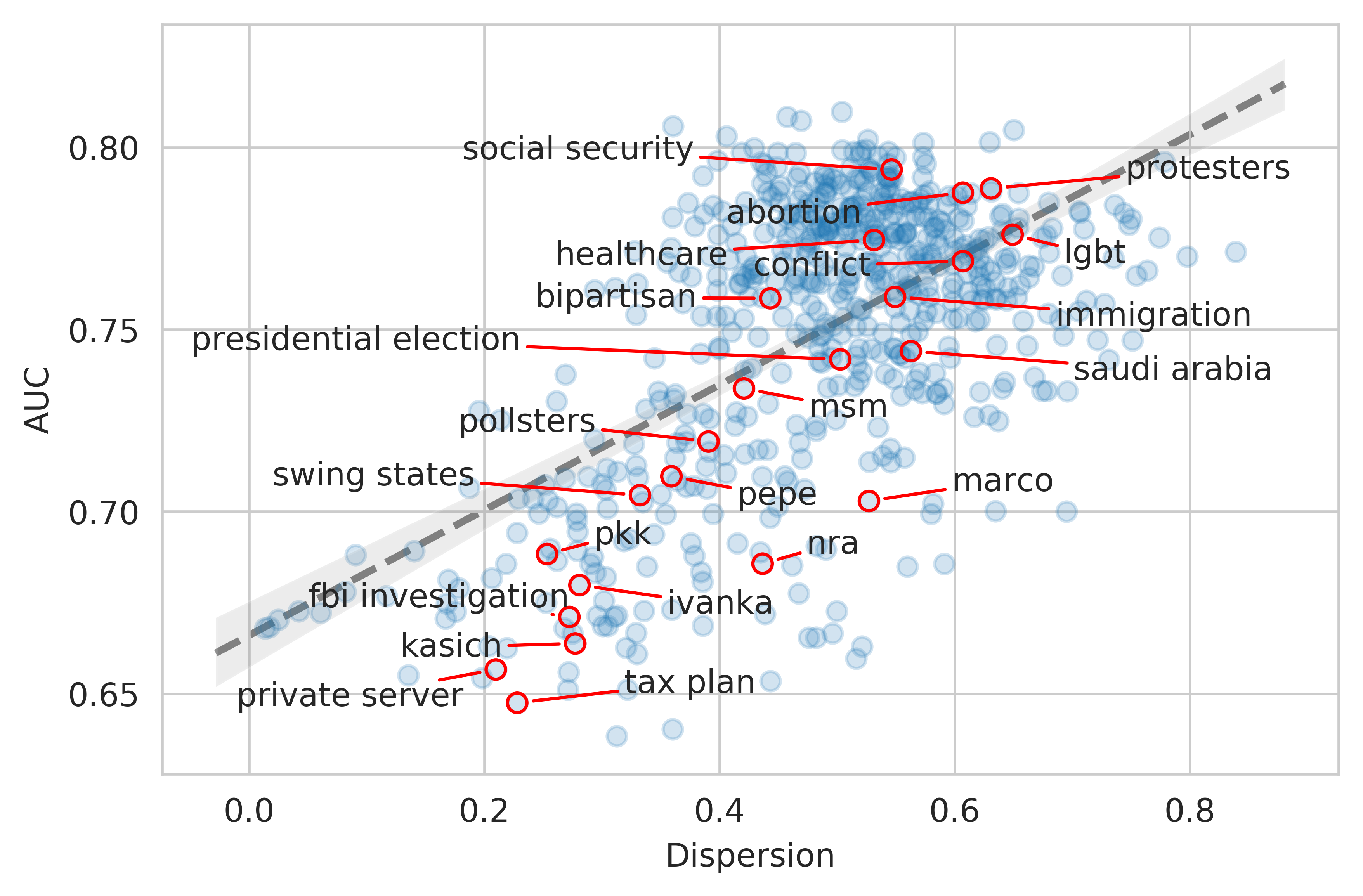}  
        \caption[]{AUC as a function of dispersion. Concepts with higher dispersion in $\mathcal{X}_*$ tend 
      to result in better performance on link prediction. Concepts with large values for dispersion and AUC are most polarized. We provide annotations for selected concepts.}
        \label{fig:reg-plot}
\end{figure}

\begin{figure}[t!]
        \centering        
        \begin{subfigure}[\small \textit{private server}]{ 
            \includegraphics[width=0.2\textwidth]{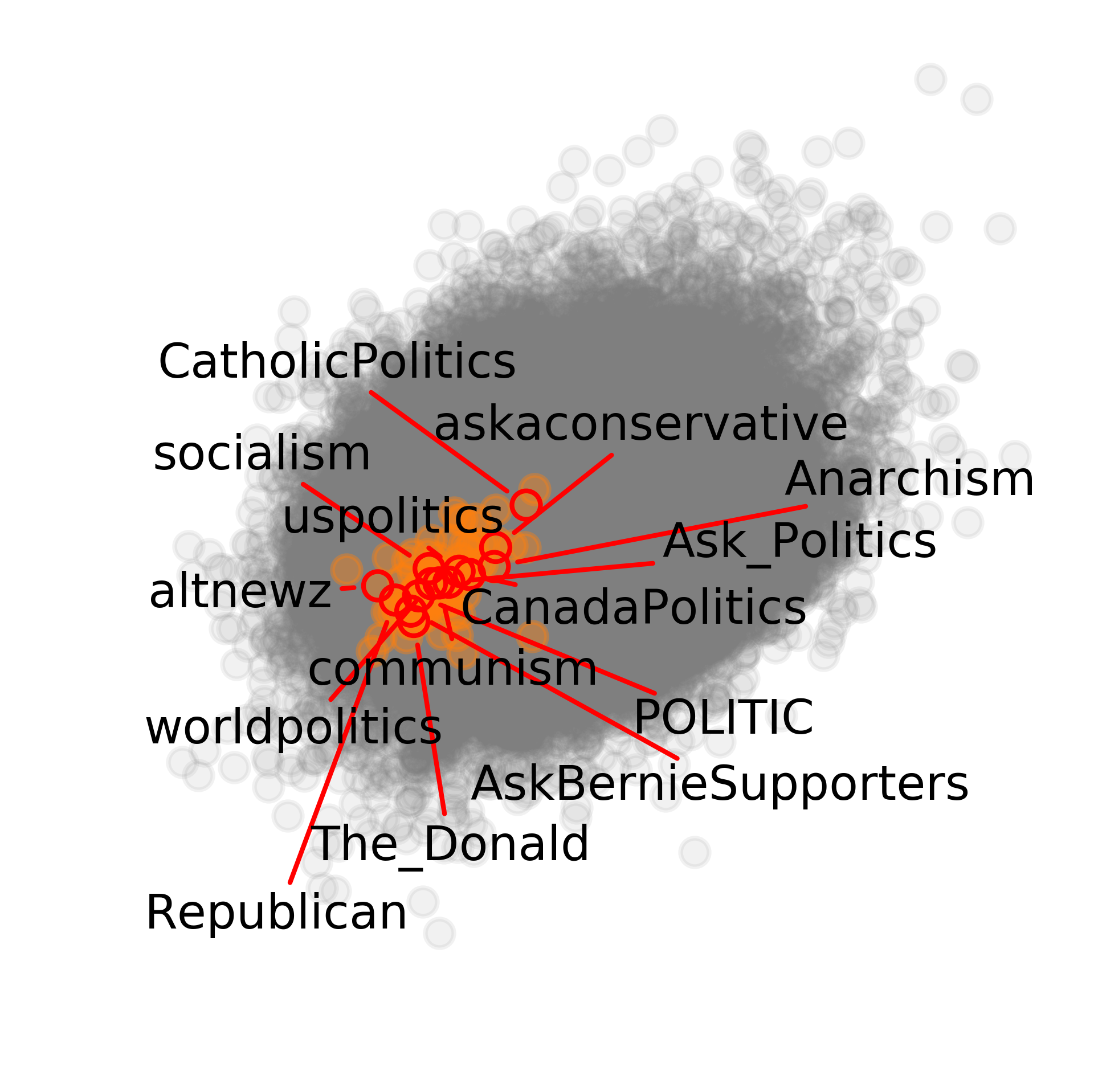}}
            \label{fig:lgbt}
        \end{subfigure}    
        \begin{subfigure}[\small \textit{lgbt}]{ 
            \includegraphics[width=0.2\textwidth]{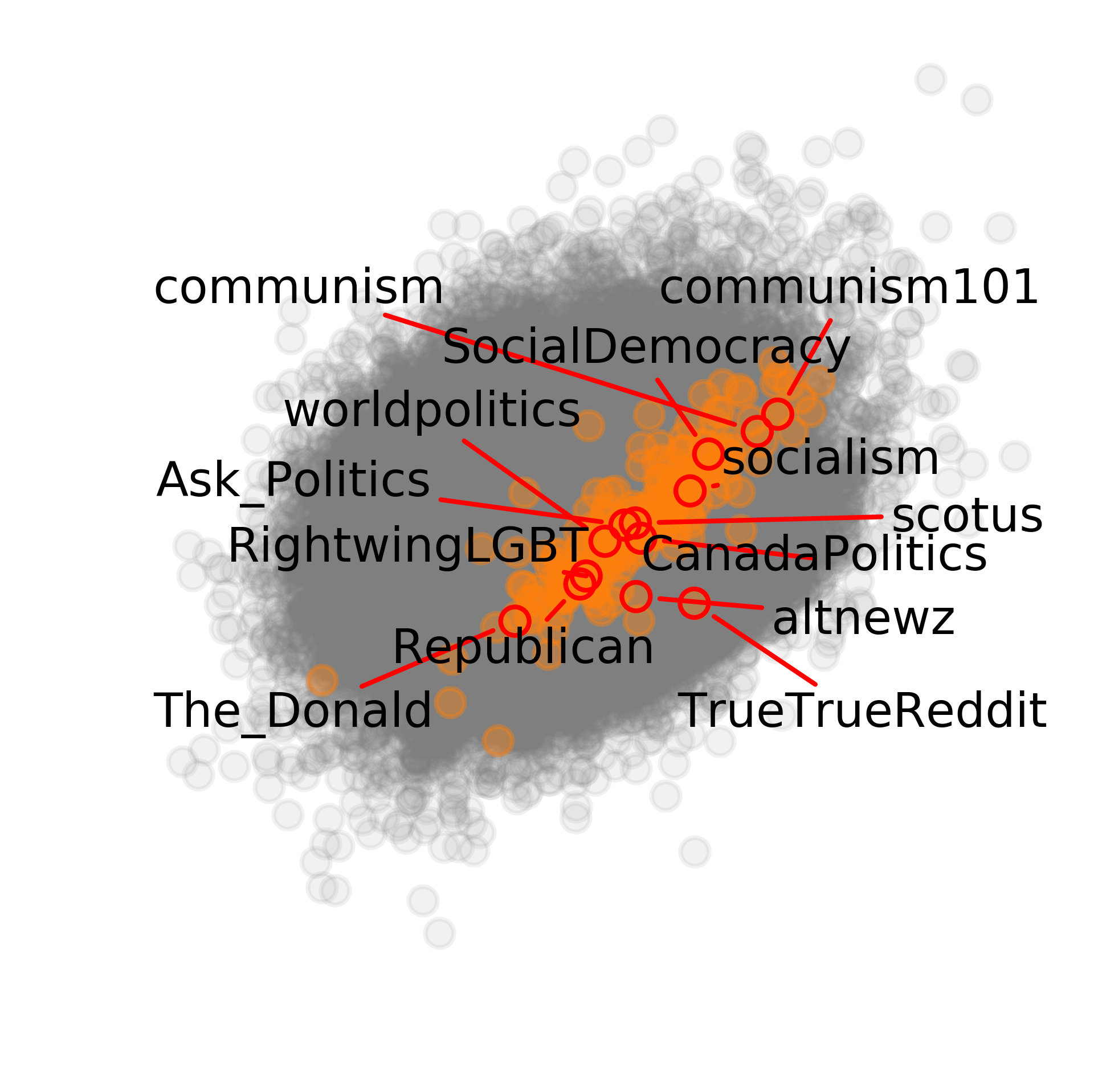}}
            \label{fig:private-server}
        \end{subfigure}
        \caption[]{Dispersion of concepts in $\mathcal{X}_*$. The gray points represent all embeddings in $\mathcal{X}_*$.  While the cluster for \textit{private server} (unpolarized) is clumped, the cluster for \textit{lgbt} (polarized) forms a skewed ellipse. We provide annotations for selected subreddits.}
        \label{fig:pca}
\end{figure}

Given our general framework, we expect concepts with more polarized 
ideological bias to (i) have
embeddings with larger dispersion in $\mathcal{X}_*$ and (ii) show
better performance on link prediction.\footnote{This section uses the 2016 data.}
Measuring dispersion 
as the average $\ell_2$ distance of a concept's embeddings in $\mathcal{X}_*$
to their centroid, we find a significant positive 
correlation between dispersion 
and AUC
($R^2=.344$, $F(1, 598) = 314$, $p<.001$). Thus,
large variance in the ideological subspace
reflects 
the (polarized) structure of the social network. Many of the concepts with large values
for both dispersion and AUC such as \textit{abortion} and \textit{lgbt} 
are known to be polarized from previous research \citep{Yardi.2010, Mendelsohn.2020}, but we also find concepts such 
as \textit{protesters} and \textit{social security} that have been studied less
(Figure~\ref{fig:reg-plot}).

Taking \textit{lgbt} (\textit{private server}) as an example for
a polarized (unpolarized) concept, we visualize 
the resulting clusters in $\mathcal{X}_*$ by means of PCA (Figure \ref{fig:pca}).
Whereas the cluster for \textit{private server} is  clumped, 
the cluster for \textit{lgbt} forms a  skewed ellipse spread across the subspace.
We further notice an ideological split for \textit{lgbt}:  left-wing and right-wing
subreddits occupy opposite ends of the ellipse.

\section{Limitations}

Instead of direct supervision (e.g., in the form of word lists), 
our method finds the bias subspace by using the assortative information 
latently encoded in the structure of social networks. We believe that 
the ubiquity and variety of social networks online makes our method more scalable 
and more
widely applicable than previous methods. However, there might be use cases for which 
high-quality external resources are readily available or social networks do not exist, and hence 
a supervised method might be preferable. 

While we only apply our method to data from the Reddit Politosphere,
the structure of Reddit as a forum divided into smaller subforums
is very common on the web and shared by some of
the most intensely researched online platforms (e.g., 4chan).
Our method can
also be applied to other types of social networks as long as
(i) they are homophilous, and (ii) they have text attached
to the nodes. For social networks whose nodes correspond to individual users (e.g., Twitter), careful preprocessing
might be required to ensure enough data per node (e.g.,
graph clustering).

The success of our method depends on how accurately 
variables relevant for bias (in this study ideology) 
are reflected by the social network, which means
that care must be taken during network selection (explicit networks) and construction (implicit networks).
For example, user overlap on Reddit can also be due to conflict between subreddits
\citep{Datta.2017, Kumar.2018, Datta.2019}. While
we do not find this to affect our results, it might
be a limitation if the degree of homophily is too low.

\section{Conclusion}

We propose
a fully unsupervised method that exploits the structure of social 
networks to detect bias in contextualized embeddings. The method combines orthogonality regularization, structured
sparsity learning, and graph neural networks. 
While we focus on the use case of ideological bias in online discussion 
forums, our method can be easily applied to other types of bias (e.g., gender bias based on
social networks encoding friendship relations or scientific bias based on citation networks).
We also present semantic and indexical probing as two 
complementary techniques to 
probe the found
subspace.
Our experiments show that the ideological subspace
encodes abstract evaluative semantics and reflects
changes in the ideological left-right spectrum during the presidency
of Donald Trump.

\section{Acknowledgements}

This work was funded by the European Research
Council (\#740516) and the Engineering and Physical
Sciences Research Council (EP/T023333/1).
The first author was also supported by the German
Academic Scholarship Foundation and the Arts
and Humanities Research Council. We thank the reviewers for
their helpful comments.

\bibliography{icml-2022}

\begin{thebibliography}{107}
\providecommand{\natexlab}[1]{#1}
\providecommand{\url}[1]{\texttt{#1}}
\expandafter\ifx\csname urlstyle\endcsname\relax
  \providecommand{\doi}[1]{doi: #1}\else
  \providecommand{\doi}{doi: \begingroup \urlstyle{rm}\Url}\fi

\bibitem[Adamic \& Glance(2005)Adamic and Glance]{Adamic.2005}
Adamic, L.~A. and Glance, N.
\newblock The political blogosphere and the 2004 {U.S.} {E}lection: Divided
  they blog.
\newblock In \emph{International Workshop on Link Discovery (LinkKDD) 3}, 2005.

\bibitem[An et~al.(2018)An, Kwak, and Ahn]{An.2018}
An, J., Kwak, H., and Ahn, Y.-Y.
\newblock Sem{A}xis: A lightweight framework to characterize domain-specific
  word semantics beyond sentiment.
\newblock In \emph{Annual Meeting of the Association for Computational
  Linguistics (ACL) 56}, 2018.

\bibitem[An et~al.(2019)An, Kwak, Posegga, and Jungherr]{An.2019}
An, J., Kwak, H., Posegga, O., and Jungherr, A.
\newblock Political discussions in homogeneous and cross-cutting communication
  spaces.
\newblock In \emph{International AAAI Conference on Web and Social Media
  (ICWSM) 13}, 2019.

\bibitem[Bach et~al.(2011)Bach, Jenatton, Mairal, and Obozinski]{Bach.2011}
Bach, F., Jenatton, R., Mairal, J., and Obozinski, G.
\newblock Optimization with sparsity-inducing penalties.
\newblock \emph{Foundations and Trends in Machine Learning}, 4\penalty0
  (1):\penalty0 1--106, 2011.

\bibitem[Bakshy et~al.(2015)Bakshy, Messing, and Adamic]{Bakshy.2015}
Bakshy, E., Messing, S., and Adamic, L.~A.
\newblock Exposure to ideologically diverse news and opinion on {F}acebook.
\newblock \emph{Science}, 384\penalty0 (6239):\penalty0 1130--1132, 2015.

\bibitem[Baly et~al.(2020)Baly, {Da San Martino}, Glass, and Nakov]{Baly.2020}
Baly, R., {Da San Martino}, G., Glass, J., and Nakov, P.
\newblock We can detect your bias: Predicting the political ideology of news
  articles.
\newblock In \emph{Conference on Empirical Methods in Natural Language
  Processing (EMNLP) 2020}, 2020.

\bibitem[Barber \& Pope(2019)Barber and Pope]{Barber.2019}
Barber, M. and Pope, J.~C.
\newblock Does party trump ideology? {D}isentangling party and ideology in
  {A}merica.
\newblock \emph{American Political Science Review}, 113\penalty0 (1):\penalty0
  38--54, 2019.

\bibitem[Basta et~al.(2019)Basta, Costa-juss{\`a}, and Casas]{Basta.2019}
Basta, C., Costa-juss{\`a}, M.~R., and Casas, N.
\newblock Evaluating the underlying gender bias in contextualized word
  embeddings.
\newblock In \emph{Workshop on Gender Bias in Natural Language Processing
  (GeBNLP) 1}, 2019.

\bibitem[Benson(2013)]{Benson.2013}
Benson, R.
\newblock \emph{Shaping immigration news: A {F}rench-{A}merican comparison}.
\newblock {Cambridge University Press}, Cambridge, UK, 2013.

\bibitem[Bianchi et~al.(2021)Bianchi, Marelli, Nicoli, and
  Palmonari]{Bianchi.2021}
Bianchi, F., Marelli, M., Nicoli, P., and Palmonari, M.
\newblock {SWEAT}: Scoring polarization of topics across different corpora.
\newblock In \emph{Conference on Empirical Methods in Natural Language
  Processing (EMNLP) 2021}, 2021.

\bibitem[Blodgett et~al.(2020)Blodgett, Barocas, {Daum{\'e} III}, and
  Wallach]{Blodgett.2020}
Blodgett, S.~L., Barocas, S., {Daum{\'e} III}, H., and Wallach, H.
\newblock Language (technology) is power: A critical survey of ``bias'' in
  {NLP}.
\newblock In \emph{Annual Meeting of the Association for Computational
  Linguistics (ACL) 58}, 2020.

\bibitem[Bolukbasi et~al.(2016)Bolukbasi, Chang, Zou, Saligrama, and
  Kalai]{Bolukbasi.2016}
Bolukbasi, T., Chang, K.-W., Zou, J., Saligrama, V., and Kalai, A.
\newblock Man is to computer programmer as woman is to homemaker? debiasing
  word embeddings.
\newblock In \emph{Advances in Neural Information Processing Systems (NIPS)
  30}, 2016.

\bibitem[Bousmalis et~al.(2016)Bousmalis, Trigeorgis, Silberman, Krishnan, and
  Erhan]{Bousmalis.2016}
Bousmalis, K., Trigeorgis, G., Silberman, N., Krishnan, D., and Erhan, D.
\newblock Domain separation networks.
\newblock In \emph{Advances in Neural Information Processing Systems (NIPS)
  30}, 2016.

\bibitem[Brock et~al.(2017)Brock, Lim, Ritchie, and Weston]{Brock.2017}
Brock, A., Lim, T., Ritchie, J.~M., and Weston, N.
\newblock Neural photo editing with introspective adversarial networks.
\newblock In \emph{International Conference on Learning Representations (ICLR)
  5}, 2017.

\bibitem[Brysbaert et~al.(2014)Brysbaert, Warriner, and
  Kuperman]{Brysbaert.2014}
Brysbaert, M., Warriner, A.~B., and Kuperman, V.
\newblock Concreteness ratings for 40 thousand generally known {E}nglish word
  lemmas.
\newblock \emph{Behavior Research Methods}, 46\penalty0 (3):\penalty0 904--911,
  2014.

\bibitem[Caliskan et~al.(2017)Caliskan, Bryson, and Narayanan]{Caliskan.2017}
Caliskan, A., Bryson, J.~J., and Narayanan, A.
\newblock Semantics derived automatically from language corpora contain
  human-like biases.
\newblock \emph{Science}, 356\penalty0 (6334):\penalty0 183--186, 2017.

\bibitem[Cann et~al.(2021)Cann, Weaver, and Williams]{Cann.2021}
Cann, T.~J., Weaver, I.~S., and Williams, H.~T.
\newblock Ideological biases in social sharing of online information about
  climate change.
\newblock \emph{PloS ONE}, 16\penalty0 (4):\penalty0 e0250656, 2021.

\bibitem[Cao et~al.(2022)Cao, Pruksachatkun, Chang, Gupta, Kumar, Dhamala, and
  Galstyan]{Cao.2022}
Cao, Y.~T., Pruksachatkun, Y., Chang, K.-W., Gupta, R., Kumar, V., Dhamala, J.,
  and Galstyan, A.
\newblock On the intrinsic and extrinsic fairness evaluation metrics for
  contextualized language representations.
\newblock In \emph{Annual Meeting of the Association for Computational
  Linguistics (ACL) 60}, 2022.

\bibitem[Carmines et~al.(2016)Carmines, Ensley, and Wagner]{Carmines.2016}
Carmines, E.~G., Ensley, M.~J., and Wagner, M.~W.
\newblock Ideological heterogeneity and the rise of {D}onald {T}rump.
\newblock \emph{The Forum}, 14\penalty0 (4):\penalty0 1631, 2016.
\newblock ISSN 2194-6183.

\bibitem[Chong \& Druckman(2007)Chong and Druckman]{Chong.2007}
Chong, D. and Druckman, J.~N.
\newblock Framing theory.
\newblock \emph{Annual Review of Political Science}, 10:\penalty0 103--126,
  2007.

\bibitem[Coenen et~al.(2019)Coenen, Reif, Yuan, Kim, Pearce, Vi{\'e}gas, and
  Wattenberg]{Coenen.2019}
Coenen, A., Reif, E., Yuan, A., Kim, B., Pearce, A., Vi{\'e}gas, F., and
  Wattenberg, M.
\newblock Visualizing and measuring the geometry of {BERT}.
\newblock In \emph{Advances in Neural Information Processing Systems (NeurIPS)
  33}, 2019.

\bibitem[Conover et~al.(2011)Conover, Ratkiewicz, Francisco, Goncalves,
  Flammini, and Menczer]{Conover.2011}
Conover, M., Ratkiewicz, J., Francisco, M., Goncalves, B., Flammini, A., and
  Menczer, F.
\newblock Political polarization on {T}witter.
\newblock In \emph{International AAAI Conference on Web and Social Media
  (ICWSM) 5}, 2011.

\bibitem[Datta \& Adar(2019)Datta and Adar]{Datta.2019}
Datta, S. and Adar, E.
\newblock Extracting inter-community conflicts in reddit.
\newblock In \emph{International AAAI Conference on Web and Social Media
  (ICWSM) 13}, 2019.

\bibitem[Datta et~al.(2017)Datta, Phelan, and Adar]{Datta.2017}
Datta, S., Phelan, C., and Adar, E.
\newblock Identifying misaligned inter-group links and communities.
\newblock \emph{Proceedings of the ACM on Human-Computer Interaction},
  1:\penalty0 1--23, 2017.

\bibitem[Davoodi et~al.(2020)Davoodi, Waltenburg, and Goldwasser]{Davoodi.2020}
Davoodi, M., Waltenburg, E., and Goldwasser, D.
\newblock Understanding the language of political agreement and disagreement in
  legislative texts.
\newblock In \emph{Annual Meeting of the Association for Computational
  Linguistics (ACL) 58}, 2020.

\bibitem[Deleu \& Bengio(2021)Deleu and Bengio]{Deleu.2021}
Deleu, T. and Bengio, Y.
\newblock Structured sparsity inducing adaptive optimizers for deep learning.
\newblock In \emph{arXiv 2102.03869}, 2021.

\bibitem[Demszky et~al.(2019)Demszky, Garg, Voigt, Zou, Gentzkow, Shapiro, and
  Jurafsky]{Demszky.2019}
Demszky, D., Garg, N., Voigt, R., Zou, J., Gentzkow, M., Shapiro, J., and
  Jurafsky, D.
\newblock Analyzing polarization in social media: Method and application to
  {T}weets on 21 mass shootings.
\newblock In \emph{Annual Conference of the North American Chapter of the
  Association for Computational Linguistics: Human Language Technologies (NAACL
  HTL) 2019}, 2019.

\bibitem[Devlin et~al.(2019)Devlin, Chang, Lee, and Toutanova]{Devlin.2019}
Devlin, J., Chang, M.-W., Lee, K., and Toutanova, K.
\newblock {BERT}: Pre-training of deep bidirectional transformers for language
  understanding.
\newblock In \emph{Annual Conference of the North American Chapter of the
  Association for Computational Linguistics: Human Language Technologies (NAACL
  HTL) 2019}, 2019.

\bibitem[DiPrete et~al.(2011)DiPrete, Gelman, McCormick, Teitler, and
  Zheng]{DiPrete.2011}
DiPrete, T.~A., Gelman, A., McCormick, T., Teitler, J., and Zheng, T.
\newblock Segregation in social networks based on acquaintanceship and trust.
\newblock \emph{American Journal of Sociology}, 116\penalty0 (4):\penalty0
  1234--1283, 2011.

\bibitem[Druckman(2001)]{Druckman.2001}
Druckman, J.~N.
\newblock The implications of framing effects for citizen competence.
\newblock \emph{Political Behavior}, 23\penalty0 (2):\penalty0 225--256, 2001.

\bibitem[Eckert(2012)]{Eckert.2012}
Eckert, P.
\newblock Three waves of variation study: The emergence of meaning in the study
  of sociolinguistic variation.
\newblock \emph{Annual Review of Anthropology}, 41\penalty0 (1):\penalty0
  87--100, 2012.

\bibitem[Eckert(2019)]{Eckert.2019}
Eckert, P.
\newblock The limits of meaning: Social indexicality, variation, and the cline
  of interiority.
\newblock \emph{Language}, 95\penalty0 (4):\penalty0 751--776, 2019.

\bibitem[Entman(1993)]{Entman.1993}
Entman, R.~M.
\newblock Framing: Toward clarification of a fractured paradigm.
\newblock \emph{Journal of Communication}, 43\penalty0 (4):\penalty0 51--58,
  1993.

\bibitem[Fan et~al.(2019)Fan, White, Sharma, Su, Choubey, Huang, and
  Wang]{Fan.2019}
Fan, L., White, M., Sharma, E., Su, R., Choubey, P.~K., Huang, R., and Wang, L.
\newblock In plain sight: Media bias through the lens of factual reporting.
\newblock In \emph{Conference on Empirical Methods in Natural Language
  Processing (EMNLP) 2019}, 2019.

\bibitem[Field \& Tsvetkov(2019)Field and Tsvetkov]{Field.2019}
Field, A. and Tsvetkov, Y.
\newblock Entity-centric contextual affective analysis.
\newblock In \emph{Annual Meeting of the Association for Computational
  Linguistics (ACL) 57}, 2019.

\bibitem[Fulgoni et~al.(2016)Fulgoni, Carpenter, Ungar, and
  Preotiuc-Pietro]{Fulgoni.2016}
Fulgoni, D., Carpenter, J., Ungar, L.~H., and Preotiuc-Pietro, D.
\newblock An empirical exploration of moral foundations theory in partisan news
  sources.
\newblock In \emph{International Conference on Language Resources and
  Evaluation (LREC) 10}, 2016.

\bibitem[Garcia et~al.(2015)Garcia, Abisheva, Schweighofer, Serd{\"u}lt, and
  Schweitzer]{Garcia.2015}
Garcia, D., Abisheva, A., Schweighofer, S., Serd{\"u}lt, U., and Schweitzer, F.
\newblock Ideological and temporal components of network polarization in online
  political participatory media.
\newblock \emph{Policy and Internet}, 7\penalty0 (1):\penalty0 46--79, 2015.

\bibitem[Garimella et~al.(2018)Garimella, Morales, Gionis, and
  Mathioudakis]{Garimella.2018}
Garimella, K., Morales, G. D.~F., Gionis, A., and Mathioudakis, M.
\newblock Quantifying controversy on social media.
\newblock \emph{ACM Transactions on Social Computing}, 1\penalty0 (1):\penalty0
  1--27, 2018.

\bibitem[Gentzkow \& Shapiro(2010)Gentzkow and Shapiro]{Gentzkow.2010}
Gentzkow, M. and Shapiro, J.~M.
\newblock What drives media slant? evidence from {U.S.} daily newspapers.
\newblock \emph{Econometrica}, 78\penalty0 (1):\penalty0 35--71, 2010.

\bibitem[Goldberg(2019)]{Goldberg.2019}
Goldberg, Y.
\newblock Assessing {BERT}'s syntactic abilities.
\newblock In \emph{arXiv 1901.05287}. 2019.

\bibitem[Gonen \& Goldberg(2019)Gonen and Goldberg]{Gonen.2019}
Gonen, H. and Goldberg, Y.
\newblock Lipstick on a pig: Debiasing methods cover up systematic gender
  biases in word embeddings but do not remove them.
\newblock In \emph{Annual Conference of the North American Chapter of the
  Association for Computational Linguistics: Human Language Technologies (NAACL
  HTL) 2019}, 2019.

\bibitem[Graham et~al.(2009)Graham, Haidt, and Nosek]{Graham.2009}
Graham, J., Haidt, J., and Nosek, B.~A.
\newblock Liberals and conservatives rely on different sets of moral
  foundations.
\newblock \emph{Journal of Personality and Social Psychology}, 96\penalty0
  (5):\penalty0 1029--1046, 2009.

\bibitem[Graham et~al.(2013)Graham, Haidt, Koleva, Motyl, Iyer, Wojcik, and
  Ditto]{Graham.2013}
Graham, J., Haidt, J., Koleva, S., Motyl, M., Iyer, R., Wojcik, S.~P., and
  Ditto, P.~H.
\newblock Moral foundations theory.
\newblock \emph{Advances in Experimental Social Psychology}, 47:\penalty0
  55--130, 2013.

\bibitem[Green et~al.(2020)Green, Edgerton, Naftel, Shoub, and
  Cranmer]{Green.2020}
Green, J., Edgerton, J., Naftel, D., Shoub, K., and Cranmer, S.
\newblock Elusive consensus: Polarization in elite communication on the
  {COVID}-19 pandemic.
\newblock \emph{Science Advances}, 6:\penalty0 eabc2717, 2020.

\bibitem[Guerra et~al.(2013)Guerra, Meira, Cardie, and Kleinberg]{Guerra.2013}
Guerra, P.~H., Meira, W., Cardie, C., and Kleinberg, R.
\newblock A measure of polarization on social media networks based on community
  boundaries.
\newblock In \emph{International AAAI Conference on Web and Social Media
  (ICWSM) 7}, 2013.

\bibitem[Guo \& Caliskan(2021)Guo and Caliskan]{Guo.2021}
Guo, W. and Caliskan, A.
\newblock Detecting emergent intersectional biases: Contextualized word
  embeddings contain a distribution of human-like biases.
\newblock In \emph{AAAI/ACM Conference on Artificial Intelligence, Ethics, and
  Society (AIES) 4}, 2021.

\bibitem[Haidt \& Graham(2007)Haidt and Graham]{Haidt.2007}
Haidt, J. and Graham, J.
\newblock When morality opposes justice: Conservatives have moral intuitions
  that liberals may not recognize.
\newblock \emph{Social Justice Research}, 20\penalty0 (1):\penalty0 98--116,
  2007.

\bibitem[Haidt \& Joseph(2004)Haidt and Joseph]{Haidt.2004}
Haidt, J. and Joseph, C.
\newblock Intuitive ethics: How innately prepared intuitions generate
  culturally variable virtues.
\newblock \emph{Daedalus}, 133\penalty0 (4):\penalty0 55--66, 2004.

\bibitem[He et~al.(2021)He, Mokhberian, C{\^a}mara, Abeliuk, and
  Lerman]{He.2021}
He, Z., Mokhberian, N., C{\^a}mara, A., Abeliuk, A., and Lerman, K.
\newblock Detecting polarized topics in {COVID}-19 news using
  partisanship-aware contextualized topic embeddings.
\newblock In \emph{Findings of the Association for Computational Linguistics:
  EMNLP 2021}, 2021.

\bibitem[Heckman \& Snyder(1997)Heckman and Snyder]{Heckman.1997}
Heckman, J.~J. and Snyder, J.~M.
\newblock Linear probability models of the demand for attributes with an
  empirical application to estimating the preferences of legislators.
\newblock \emph{The RAND Journal of Economics}, 28\penalty0 (0):\penalty0
  142--189, 1997.

\bibitem[Hewitt \& Manning(2019)Hewitt and Manning]{Hewitt.2019}
Hewitt, J. and Manning, C.~D.
\newblock A structural probe for finding syntax in word representations.
\newblock In \emph{Annual Conference of the North American Chapter of the
  Association for Computational Linguistics: Human Language Technologies (NAACL
  HTL) 2019}, 2019.

\bibitem[Heywood(2017)]{Heywood.2017}
Heywood, A.
\newblock \emph{Political ideologies: An introduction}.
\newblock Macmillan, London, UK, 2017.

\bibitem[Himelboim et~al.(2013)Himelboim, McCreery, and Smith]{Himelboim.2013}
Himelboim, I., McCreery, S., and Smith, M.
\newblock Birds of a feather tweet together: Integrating network and content
  analyses to examine cross-ideology exposure on {T}witter.
\newblock \emph{Journal of Computer-Mediated Communication}, 18\penalty0
  (2):\penalty0 40--60, 2013.

\bibitem[Hofmann et~al.(2022{\natexlab{a}})Hofmann, Dong, Pierrehumbert, and
  Sch{\"u}tze]{Hofmann.2022b}
Hofmann, V., Dong, X., Pierrehumbert, J.~B., and Sch{\"u}tze, H.
\newblock Modeling ideological salience and framing in polarized online groups
  with graph neural networks and structured sparsity.
\newblock In \emph{Findings of the Association for Computational Linguistics:
  NAACL 2022}, 2022{\natexlab{a}}.

\bibitem[Hofmann et~al.(2022{\natexlab{b}})Hofmann, Sch{\"u}tze, and
  Pierrehumbert]{Hofmann.2022}
Hofmann, V., Sch{\"u}tze, H., and Pierrehumbert, J.~B.
\newblock The {R}eddit politosphere: A large-scale text and network resource of
  online political discourse.
\newblock In \emph{International AAAI Conference on Web and Social Media
  (ICWSM) 16}, 2022{\natexlab{b}}.

\bibitem[Hopp et~al.(2021)Hopp, Fisher, Cornell, Huskey, and Weber]{Hopp.2021}
Hopp, F.~R., Fisher, J.~T., Cornell, D., Huskey, R., and Weber, R.
\newblock The extended {M}oral {F}oundations {D}ictionary (e{MFD}): Development
  and applications of a crowd-sourced approach to extracting moral intuitions
  from text.
\newblock \emph{Behavior Research Methods}, 53\penalty0 (1):\penalty0 232--246,
  2021.

\bibitem[Iyyer et~al.(2014)Iyyer, Enns, Boyd-Graber, and Resnik]{Iyyer.2014}
Iyyer, M., Enns, P., Boyd-Graber, J., and Resnik, P.
\newblock Political ideology detection using recursive neural networks.
\newblock In \emph{Annual Meeting of the Association for Computational
  Linguistics (ACL) 52}, 2014.

\bibitem[Jiang \& Fellbaum(2020)Jiang and Fellbaum]{Jiang.2020b}
Jiang, M. and Fellbaum, C.
\newblock Interdependencies of gender and race in contextualized word
  embeddings.
\newblock In \emph{Workshop on Gender Bias in Natural Language Processing
  (GeBNLP) 2}, 2020.

\bibitem[Kingma \& Ba(2015)Kingma and Ba]{Kingma.2015}
Kingma, D.~P. and Ba, J.~L.
\newblock Adam: A method for stochastic optimization.
\newblock In \emph{International Conference on Learning Representations (ICLR)
  3}, 2015.

\bibitem[Kipf \& Welling(2016)Kipf and Welling]{Kipf.2016}
Kipf, T.~N. and Welling, M.
\newblock Variational graph auto-encoders.
\newblock In \emph{NIPS Bayesian Deep Learning Workshop}, 2016.

\bibitem[Kipf \& Welling(2017)Kipf and Welling]{Kipf.2017}
Kipf, T.~N. and Welling, M.
\newblock Semi-supervised classification with graph convolutional networks.
\newblock In \emph{International Conference on Learning Representations (ICLR)
  5}, 2017.

\bibitem[Knoche et~al.(2019)Knoche, Popovi{\'c}, Lemmerich, and
  Strohmaier]{Knoche.2019}
Knoche, M., Popovi{\'c}, R., Lemmerich, F., and Strohmaier, M.
\newblock Identifying biases in politically biased {Wikis} through word
  embeddings.
\newblock In \emph{ACM Conference on Hypertext and Social Media (HT) 30}, 2019.

\bibitem[Kulkarni et~al.(2018)Kulkarni, Ye, Skiena, and Wang]{Kulkarni.2018}
Kulkarni, V., Ye, J., Skiena, S., and Wang, W.~Y.
\newblock Multi-view models for political ideology detection of news articles.
\newblock In \emph{Conference on Empirical Methods in Natural Language
  Processing (EMNLP) 2018}, 2018.

\bibitem[Kumar et~al.(2018)Kumar, Hamilton, Leskovec, and Jurafsky]{Kumar.2018}
Kumar, S., Hamilton, W., Leskovec, J., and Jurafsky, D.
\newblock Community interaction and conflict on the web.
\newblock In \emph{The Web Conference (WWW) 27}, 2018.

\bibitem[Lebedev \& Lempitsky(2016)Lebedev and Lempitsky]{Lebedev.2016}
Lebedev, V. and Lempitsky, V.
\newblock Fast {C}onv{N}ets using group-wise brain damage.
\newblock In \emph{Conference on Computer Vision and Pattern Recognition (CVPR)
  29}, 2016.

\bibitem[Liu et~al.(2015)Liu, Wang, Foroosh, Tappen, and Pensky]{Liu.2015}
Liu, B., Wang, M., Foroosh, H., Tappen, M., and Pensky, M.
\newblock Sparse convolutional neural networks.
\newblock In \emph{Conference on Computer Vision and Pattern Recognition (CVPR)
  28}, 2015.

\bibitem[Massachs et~al.(2020)Massachs, Monti, Morales, and
  Bonchi]{Massachs.2020}
Massachs, J., Monti, C., Morales, G. D.~F., and Bonchi, F.
\newblock Roots of {T}rumpism: Homophily and social feedback in {D}onald
  {T}rump support on {R}eddit.
\newblock In \emph{ACM Conference on Web Science (WebSci) 12}, 2020.

\bibitem[McNemar(1947)]{McNemar.1947}
McNemar, Q.
\newblock Note on the sampling error of the difference between correlated
  proportions or percentages.
\newblock \emph{Psychometrika}, 12\penalty0 (2):\penalty0 153--157, 1947.

\bibitem[McPherson et~al.(2001)McPherson, Smith-Lovin, and
  Cook]{McPherson.2001}
McPherson, M., Smith-Lovin, L., and Cook, J.~M.
\newblock Birds of a feather: Homophily in social networks.
\newblock \emph{Annual Review of Sociology}, 27:\penalty0 415--444, 2001.

\bibitem[Mejova et~al.(2014)Mejova, Zhang, Diakopoulos, and
  Castillo]{Mejova.2014}
Mejova, Y., Zhang, A.~X., Diakopoulos, N., and Castillo, C.
\newblock Controversy and sentiment in online news.
\newblock In \emph{arXiv 1409.8152}. 2014.

\bibitem[Mendelsohn et~al.(2020)Mendelsohn, Tsvetkov, and
  Jurafsky]{Mendelsohn.2020}
Mendelsohn, J., Tsvetkov, Y., and Jurafsky, D.
\newblock A framework for the computational linguistic analysis of
  dehumanization.
\newblock \emph{Frontiers in Artificial Intelligence}, 3:\penalty0 55, 2020.

\bibitem[Mendelsohn et~al.(2021)Mendelsohn, Budak, and
  Jurgens]{Mendelsohn.2021}
Mendelsohn, J., Budak, C., and Jurgens, D.
\newblock Modeling framing in immigration discourse on social media.
\newblock In \emph{Annual Conference of the North American Chapter of the
  Association for Computational Linguistics: Human Language Technologies (NAACL
  HTL) 2021}, 2021.

\bibitem[Mokhberian et~al.(2020)Mokhberian, Abeliuk, Cummings, and
  Lerman]{Mokhberian.2020}
Mokhberian, N., Abeliuk, A., Cummings, P., and Lerman, K.
\newblock Moral framing and ideological bias of news.
\newblock In \emph{International Conference on Social Informatics (SocInfo)
  12}, 2020.

\bibitem[Nelson et~al.(1997)Nelson, Oxley, and Clawson]{Nelson.1997}
Nelson, T.~E., Oxley, Z.~M., and Clawson, R.~A.
\newblock Toward a psychology of framing effects.
\newblock \emph{Political Behavior}, 19\penalty0 (3):\penalty0 221--246, 1997.

\bibitem[Nguyen et~al.(2021)Nguyen, Rosseel, and Grieve]{Nguyen.2021}
Nguyen, D., Rosseel, L., and Grieve, J.
\newblock On learning and representing social meaning in {NLP}: A
  sociolinguistic perspective.
\newblock In \emph{Annual Conference of the North American Chapter of the
  Association for Computational Linguistics: Human Language Technologies (NAACL
  HTL) 2021}, 2021.

\bibitem[Nguyen et~al.(2016)Nguyen, {Schulte im Walde}, and Vu]{Nguyen.2016b}
Nguyen, K.~A., {Schulte im Walde}, S., and Vu, N.~T.
\newblock Integrating distributional lexical contrast into word embeddings for
  antonym-synonym distinction.
\newblock In \emph{Annual Meeting of the Association for Computational
  Linguistics (ACL) 54}, 2016.

\bibitem[Nguyen et~al.(2017)Nguyen, {Schulte im Walde}, and Vu]{Nguyen.2017}
Nguyen, K.~A., {Schulte im Walde}, S., and Vu, N.~T.
\newblock Distinguishing antonyms and synonyms in a pattern-based neural
  network.
\newblock In \emph{Conference of the European Chapter of the Association for
  Computational Linguistics (EACL) 15}, 2017.

\bibitem[Olson \& Neal(2015)Olson and Neal]{Olson.2015}
Olson, R.~S. and Neal, Z.~P.
\newblock Navigating the massive world of {R}eddit: Using backbone networks to
  map user interests in social media.
\newblock \emph{PeerJ Computer Science}, 2015.

\bibitem[Parikh \& Boyd(2013)Parikh and Boyd]{Parikh.2013}
Parikh, N. and Boyd, S.
\newblock Proximal algorithms.
\newblock \emph{Foundations and Trends in Optimization}, 1\penalty0
  (3):\penalty0 123--231, 2013.

\bibitem[Preotiuc-Pietro et~al.(2017)Preotiuc-Pietro, Liu, Hopkins, and
  Ungar]{PreotiucPietro.2017}
Preotiuc-Pietro, D., Liu, Y., Hopkins, D.~J., and Ungar, L.~H.
\newblock Beyond binary labels: Political ideology prediction of {T}witter
  users.
\newblock In \emph{Annual Meeting of the Association for Computational
  Linguistics (ACL) 55}, 2017.

\bibitem[Psylla et~al.(2017)Psylla, Sapiezynski, Mones, and
  Lehmann]{Psylla.2017}
Psylla, I., Sapiezynski, P., Mones, E., and Lehmann, S.
\newblock The role of gender in social network organization.
\newblock \emph{PloS ONE}, 12\penalty0 (12):\penalty0 e0189873, 2017.

\bibitem[Roy \& Goldwasser(2020)Roy and Goldwasser]{Roy.2020}
Roy, S. and Goldwasser, D.
\newblock Weakly supervised learning of nuanced frames for analyzing
  polarization in news media.
\newblock In \emph{Conference on Empirical Methods in Natural Language
  Processing (EMNLP) 2020}, 2020.

\bibitem[Rozado \& al~Gharbi(2021)Rozado and al~Gharbi]{Rozado.2021}
Rozado, D. and al~Gharbi, M.
\newblock Using word embeddings to probe sentiment associations of politically
  loaded terms in news and opinion articles from news media outlets.
\newblock \emph{Journal of Computational Social Science}, 56\penalty0
  (3):\penalty0 256, 2021.

\bibitem[Sagi et~al.(2013)Sagi, Diermeier, and Kaufmann]{Sagi.2013}
Sagi, E., Diermeier, D., and Kaufmann, S.
\newblock Identifying issue frames in text.
\newblock \emph{PloS ONE}, 8\penalty0 (7):\penalty0 e69185, 2013.

\bibitem[Satop{\"a}{\"a} et~al.(2011)Satop{\"a}{\"a}, Albrecht, Irwin, and
  Raghavan]{Satopaa.2011}
Satop{\"a}{\"a}, V., Albrecht, J., Irwin, D., and Raghavan, B.
\newblock Finding a ``kneedle'' in a haystack: Detecting knee points in system
  behavior.
\newblock In \emph{International Conference on Distributed Computing Systems
  (ICDCS) 31}, 2011.

\bibitem[Shen \& Ros{\'e}(2019)Shen and Ros{\'e}]{Shen.2019}
Shen, Q. and Ros{\'e}, C.
\newblock The discourse of online content moderation: Investigating polarized
  user responses to changes in {R}eddit's quarantine policy.
\newblock In \emph{Workshop on Abusive Language Online 3}, 2019.

\bibitem[Shwartz et~al.(2017)Shwartz, Santus, and Schlechtweg]{Shwartz.2017}
Shwartz, V., Santus, E., and Schlechtweg, D.
\newblock Hypernyms under siege: Linguistically-motivated artillery for
  hypernymy detection.
\newblock In \emph{Conference of the European Chapter of the Association for
  Computational Linguistics (EACL) 15}, 2017.

\bibitem[Silverstein(2003)]{Silverstein.2003}
Silverstein, M.
\newblock Indexical order and the dialectics of sociolinguistic life.
\newblock \emph{Language {\&} Communication}, 23\penalty0 (3-4):\penalty0
  193--229, 2003.

\bibitem[Sylwester \& Purver(2015)Sylwester and Purver]{Sylwester.2015}
Sylwester, K. and Purver, M.
\newblock Twitter language use reflects psychological differences between
  democrats and republicans.
\newblock \emph{PloS ONE}, pp.\  e0137422, 2015.

\bibitem[Tan \& Celis(2019)Tan and Celis]{Tan.2019}
Tan, Y.~C. and Celis, L.~E.
\newblock Assessing social and intersectional biases in contextualized word
  representations.
\newblock In \emph{Advances in Neural Information Processing Systems (NeurIPS)
  33}, 2019.

\bibitem[Tripodi et~al.(2019)Tripodi, Warglien, Sullam, and Paci]{Tripodi.2019}
Tripodi, R., Warglien, M., Sullam, S.~L., and Paci, D.
\newblock Tracing antisemitic language through diachronic embedding
  projections: {F}rance 1789-1914.
\newblock In \emph{International Workshop on Computational Approaches to
  Historical Language Change 1}, 2019.

\bibitem[Tyagi et~al.(2020)Tyagi, Field, Lathwal, Tsvetkov, and
  Carley]{Tyagi.2020}
Tyagi, A., Field, A., Lathwal, P., Tsvetkov, Y., and Carley, K.~M.
\newblock A computational analysis of polarization on {I}ndian and {P}akistani
  social media.
\newblock In \emph{International Conference on Social Informatics (SocInfo)
  12}, 2020.

\bibitem[Vargas \& Cotterell(2020)Vargas and Cotterell]{Vargas.2020}
Vargas, F. and Cotterell, R.
\newblock Exploring the linear subspace hypothesis in gender bias mitigation.
\newblock In \emph{Conference on Empirical Methods in Natural Language
  Processing (EMNLP) 2020}, 2020.

\bibitem[Vorakitphan et~al.(2020)Vorakitphan, Guerini, Cabrio, and
  Villata]{Vorakitphan.2020}
Vorakitphan, V., Guerini, M., Cabrio, E., and Villata, S.
\newblock Regrexit or not regrexit: Aspect-based sentiment analysis in
  polarized contexts.
\newblock In \emph{International Conference on Computational Linguistics
  (COLING) 28}, 2020.

\bibitem[Vorontsov et~al.(2017)Vorontsov, Trabelsi, Kadoury, and
  Pal]{Vorontsov.2017}
Vorontsov, E., Trabelsi, C., Kadoury, S., and Pal, C.
\newblock On orthogonality and learning recurrent networks with long term
  dependencies.
\newblock In \emph{International Conference on Machine Learning (ICML) 34},
  2017.

\bibitem[Waller \& Anderson(2021)Waller and Anderson]{Waller.2021}
Waller, I. and Anderson, A.
\newblock Quantifying social organization and political polarization in online
  platforms.
\newblock \emph{Nature}, 600\penalty0 (7888):\penalty0 264--268, 2021.

\bibitem[Walter et~al.(2021)Walter, Kirschner, Eger, Glava\v{s}, Lauscher, and
  Ponzetto]{Walter.2021}
Walter, T., Kirschner, C., Eger, S., Glava\v{s}, G., Lauscher, A., and
  Ponzetto, S.~P.
\newblock Diachronic analysis of {G}erman parliamentary proceedings:
  Ideological shifts through the lens of political biases.
\newblock In \emph{arXiv 2108.06295}. 2021.

\bibitem[Warriner et~al.(2013)Warriner, Kuperman, and Brysbaert]{Warriner.2013}
Warriner, A., Kuperman, V., and Brysbaert, M.
\newblock Norms of valence, arousal, and dominance for 13,915 english lemmas.
\newblock \emph{Behavior Research Methods}, 45\penalty0 (4):\penalty0
  1191--1207, 2013.

\bibitem[Weber et~al.(2013)Weber, Garimella, and Batayneh]{Weber.2013}
Weber, I., Garimella, K., and Batayneh, A.
\newblock Secular vs. islamist polarization in {E}gypt on {T}witter.
\newblock In \emph{International Conference on Advances in Social Networks
  Analysis and Mining (ASONAM) 2013}, 2013.

\bibitem[Webson et~al.(2020)Webson, Chen, Eickhoff, and Pavlick]{Webson.2020}
Webson, A., Chen, Z., Eickhoff, C., and Pavlick, E.
\newblock Do ``undocumented workers'' == ``illegal aliens''? {D}ifferentiating
  denotation and connotation in vector spaces.
\newblock In \emph{Conference on Empirical Methods in Natural Language
  Processing (EMNLP) 2020}, 2020.

\bibitem[Wen et~al.(2016)Wen, Wu, Wang, Chen, and Li]{Wen.2016}
Wen, W., Wu, C., Wang, Y., Chen, Y., and Li, H.
\newblock Learning structured sparsity in deep neural networks.
\newblock In \emph{Advances in Neural Information Processing Systems (NIPS)
  30}, 2016.

\bibitem[Wiedemann et~al.(2019)Wiedemann, Remus, Chawla, and
  Biemann]{Wiedemann.2019}
Wiedemann, G., Remus, S., Chawla, A., and Biemann, C.
\newblock Does {BERT} make any sense? interpretable word sense disambiguation
  with contextualized embeddings.
\newblock In \emph{arXiv 1909.10430}. 2019.

\bibitem[Xie et~al.(2019)Xie, {Ferreira Pinto}, Hirst, and Xu]{Xie.2019}
Xie, J.~Y., {Ferreira Pinto}, R., Hirst, G., and Xu, Y.
\newblock Text-based inference of moral sentiment change.
\newblock In \emph{Conference on Empirical Methods in Natural Language
  Processing (EMNLP) 2019}, 2019.

\bibitem[Yardi \& Boyd(2010)Yardi and Boyd]{Yardi.2010}
Yardi, S. and Boyd, D.
\newblock Dynamic debates: An analysis of group polarization over time on
  {T}witter.
\newblock \emph{Bulletin of Science, Technology {\&} Society}, 30\penalty0
  (5):\penalty0 316--327, 2010.

\bibitem[Yoon \& Hwang(2017)Yoon and Hwang]{Yoon.2017}
Yoon, J. and Hwang, S.~J.
\newblock Combined group and exclusive sparsity for deep neural networks.
\newblock In \emph{International Conference on Machine Learning (ICML) 34},
  2017.

\bibitem[Yuan \& Lin(2006)Yuan and Lin]{Yuan.2006}
Yuan, M. and Lin, Y.
\newblock Model selection and estimation in regression with grouped variables.
\newblock \emph{Journal of the Royal Statistical Society}, 68\penalty0
  (1):\penalty0 49--67, 2006.

\bibitem[Zhao et~al.(2019)Zhao, Wang, Yatskar, Cotterell, Ordonez, and
  Chang]{Zhao.2019}
Zhao, J., Wang, T., Yatskar, M., Cotterell, R., Ordonez, V., and Chang, K.-W.
\newblock Gender bias in contextualized word embeddings.
\newblock In \emph{Annual Conference of the North American Chapter of the
  Association for Computational Linguistics: Human Language Technologies (NAACL
  HTL) 2019}, 2019.

\end{thebibliography}
\bibliographystyle{icml2022}

\appendix
\section{Appendices}
\label{sec:appendix}

\subsection{Hyperparameters and Training Details} \label{ap:setup}

\begin{table} [t!]
\caption{Training statistics. $\mu_m$, $\sigma_m$: mean and standard deviation of MAUC performance on dev 
for all hyperparameter search trials; $r$: learning rate; $\lambda_o$: orthogonality constant (only $\mathcal{X}_*$); $\lambda_s$: sparsity constant (only $\mathcal{X}_*$); $\tau$: runtime in seconds for full hyperparameter search.}  \label{tab:training-stats}
\centering
\resizebox{\linewidth}{!}{%
\begin{tabular}{@{}llrrrrrr@{}}
\toprule
Space & Year & $\mu_m$ & $\sigma_m$ &  $r$ &  $\lambda_o$  & $\lambda_s$ & $\tau$\\
\midrule
\multirow{7}{*}{$\mathcal{X}_*$} 
& 2013 & .619 & .021 & 1e-04 & 3e-03 & 1e-02 & 39,633 \\
& 2014 & .599 & .045 & 3e-04 & 1e-03 & 1e-02 & 40,489 \\
& 2015 & .625 & .016 & 3e-04 & 1e-03 & 1e-02 & 40,757 \\
& 2016 & .664 & .016 & 3e-04 & 1e-03 & 1e-02 & 43,315 \\
& 2017 & .662 & .025 & 1e-04 & 1e-03 & 3e-02 & 44,341 \\
& 2018 & .636 & .015 & 1e-04 & 1e-03 & 1e-02 & 44,665 \\
& 2019 & .705 & .019 & 1e-04 & 1e-03 & 3e-02 & 47,159 \\
\midrule
\multirow{7}{*}{$\mathcal{X}$} 
& 2013 & .623 & .024 & 1e-04 & --- & --- & 3,297 \\
& 2014 & .651 & .032 & 1e-03 & --- & --- & 3,333 \\
& 2015 & .648 & .024 & 3e-04 & --- & --- & 3,335 \\
& 2016 & .665 & .030 & 3e-04 & --- & --- & 3,543 \\
& 2017 & .674 & .021 & 1e-04 & --- & --- & 3,553 \\
& 2018 & .655 & .016 & 1e-04 & --- & --- & 3,536 \\
& 2019 & .721 & .019 & 3e-04 & --- & --- & 3,717 \\
\bottomrule
\end{tabular}}
\end{table}

Both hidden layers of the graph auto-encoder have 10 dimensions. The number of trainable parameters 
is 7,800 ($\mathcal{X}$) and 597,624 ($\mathcal{X}_*$), with the latter
shrinking during training as a result of the sparsity penalty. 

Table~\ref{tab:training-stats} provides training statistics such as  
mean and standard deviation of the MAUC performance on dev,
best hyperparameter configurations, and runtimes. Experiments are performed on a GeForce GTX 1080 Ti GPU (11GB).

\subsection{Nominal and Verbal Semantic Axes} \label{ap:axes}

Nominal and verbal semantic axes with highest and lowest $s_a$ scores are provided in Tables \ref{tab:nom-axes} and 
\ref{tab:verb-axes}.
Similar to adjectives,
while top axes tend to have abstract evaluative meanings, this is 
not the case for bottom axes. 

\newpage

\begin{table} [htp!]
\caption{Top and bottom nominal semantic axes. For each year, the table shows the four nominal semantic axes with 
highest and lowest $s_a$ scores.}  \label{tab:nom-axes}
\centering
\resizebox{\linewidth}{!}{%
\begin{tabular}{@{}llllllllllllll}
\toprule
Year & Top & Bottom \\
\midrule
\multirow{4}{*} {2013} & \textit{assets/liabilities} & \textit{dislike/liking}\\ 
 & \textit{objective/subjective} & \textit{dwarf/giant}\\ 
 & \textit{divestment/investment} & \textit{nay/yea}\\ 
 & \textit{immorality/morality} & \textit{husband/wife}\\ 
\midrule 
\multirow{4}{*} {2014} & \textit{patriot/traitor} & \textit{inside/outside}\\ 
 & \textit{citizen/foreigner} & \textit{permanent/temporary}\\ 
 & \textit{impossibility/possibility} & \textit{higher/lower}\\ 
 & \textit{ability/inability} & \textit{external/internal}\\ 
\midrule 
\multirow{4}{*} {2015} & \textit{assets/liabilities} & \textit{major/minor}\\ 
 & \textit{objective/subjective} & \textit{king/queen}\\ 
 & \textit{belief/skepticism} & \textit{defeat/victory}\\ 
 & \textit{demand/supply} & \textit{decision/knockout}\\ 
\midrule 
\multirow{4}{*} {2016} & \textit{impossibility/possibility} & \textit{inside/outside}\\ 
 & \textit{credit/debit} & \textit{closing/opening}\\ 
 & \textit{guilt/innocence} & \textit{comedy/drama}\\ 
 & \textit{freeman/slave} & \textit{east/west}\\ 
\midrule 
\multirow{4}{*} {2017} & \textit{evolution/revolution} & \textit{husband/wife}\\ 
 & \textit{analysis/synthesis} & \textit{afternoon/morning}\\ 
 & \textit{barbarism/culture} & \textit{closing/opening}\\ 
 & \textit{objective/subjective} & \textit{tomorrow/yesterday}\\ 
\midrule 
\multirow{4}{*} {2018} & \textit{anarchy/government} & \textit{afternoon/morning}\\ 
 & \textit{barbarism/culture} & \textit{known/unknown}\\ 
 & \textit{deflation/inflation} & \textit{higher/lower}\\ 
 & \textit{immorality/morality} & \textit{tomorrow/yesterday}\\ 
\midrule 
\multirow{4}{*} {2019} & \textit{immorality/morality} & \textit{everybody/nobody}\\ 
 & \textit{barbarism/culture} & \textit{afternoon/morning}\\ 
 & \textit{deflation/inflation} & \textit{standing/working}\\ 
 & \textit{client/server} & \textit{anything/nothing}\\ 
\bottomrule
\end{tabular}}
\end{table}

\newpage

\begin{table} [htp!]
\caption{Top and bottom verbal semantic axes. For each year, the table shows the four verbal semantic axes with 
highest and lowest $s_a$ scores.}  \label{tab:verb-axes}
\centering
\resizebox{\linewidth}{!}{%
\begin{tabular}{@{}lllllllllll}
\toprule
Year & Top & Bottom \\
\midrule
\multirow{4}{*} {2013} & \textit{criminalize/decriminalize} & \textit{acknowledge/deny}\\ 
 & \textit{ameliorate/exacerbate} & \textit{hate/love}\\ 
 & \textit{plummet/skyrocket} & \textit{shout/whisper}\\ 
 & \textit{complicate/simplify} & \textit{bless/damn}\\ 
\midrule 
\multirow{4}{*} {2014} & \textit{plummet/skyrocket} & \textit{prefix/suffix}\\ 
 & \textit{deport/repatriate} & \textit{acknowledge/deny}\\ 
 & \textit{emigrate/immigrate} & \textit{avoid/confront}\\ 
 & \textit{disqualify/qualify} & \textit{couple/decouple}\\ 
\midrule 
\multirow{4}{*} {2015} & \textit{agree/disagree} & \textit{forget/remember}\\ 
 & \textit{decrypt/encrypt} & \textit{cease/continue}\\ 
 & \textit{conform/deviate} & \textit{get/miss}\\ 
 & \textit{plummet/skyrocket} & \textit{move/stay}\\ 
\midrule 
\multirow{4}{*} {2016} & \textit{criminalize/decriminalize} & \textit{guess/know}\\ 
 & \textit{entice/frighten} & \textit{lock/unlock}\\ 
 & \textit{plummet/skyrocket} & \textit{enter/exit}\\ 
 & \textit{hasten/postpone} & \textit{head/tail}\\ 
\midrule 
\multirow{4}{*} {2017} & \textit{centralize/decentralize} & \textit{lock/unlock}\\ 
 & \textit{generalize/specialize} & \textit{irritate/please}\\ 
 & \textit{dehumanize/humanize} & \textit{relax/stress}\\ 
 & \textit{overpay/underpay} & \textit{hate/love}\\ 
\midrule 
\multirow{4}{*} {2018} & \textit{deport/repatriate} & \textit{relax/stress}\\ 
 & \textit{criminalize/decriminalize} & \textit{guess/know}\\ 
 & \textit{centralize/decentralize} & \textit{forget/remember}\\ 
 & \textit{detain/liberate} & \textit{keep/let}\\ 
\midrule 
\multirow{4}{*} {2019} & \textit{entice/frighten} & \textit{guess/know}\\ 
 & \textit{generalize/specialize} & \textit{relax/stress}\\ 
 & \textit{centralize/decentralize} & \textit{agree/disagree}\\ 
 & \textit{elevate/relegate} & \textit{forget/remember}\\  
\bottomrule
\end{tabular}}
\end{table}

\end{document}